\pdfoutput=1
\documentclass[11pt]{article}

\newif\ifarxiv
\arxivtrue

\ifarxiv
\usepackage{acl}
\else
\usepackage[review]{acl}
\fi

\usepackage{times}
\usepackage{latexsym}

\usepackage[T1]{fontenc}
\usepackage[utf8]{inputenc}

\usepackage{microtype}

\usepackage{inconsolata}

\usepackage{graphicx}

\usepackage{todonotes}
\usepackage{booktabs}
\usepackage{multirow}
\usepackage{caption}
\usepackage{subcaption}
\usepackage{longtable}
\usepackage{tabularx}
\usepackage{longtable}
\usepackage{arydshln}
\usepackage{enumitem}
\usepackage{color}
\usepackage{hyperref}
\usepackage{soul}
\usepackage{booktabs}   % For \toprule, \midrule, \bottomrule, \cmidrule
\usepackage{siunitx}    % For S-type columns (decimal alignment)

\newcommand*\samethanks[1][\value{footnote}]{\footnotemark[#1]}
\usepackage{amsmath}

\title{
Revealing Hidden Mechanisms of Cross-Country Content Moderation
with Natural Language Processing
}

\author{
  Neemesh Yadav\thanks{Equal contribution} \\
  IIIT Delhi \\
  \texttt{neemesh20529@iiitd.ac.in} \\\And
  Jiarui Liu\samethanks \\
  CMU \\
  \texttt{jiarui@cmu.edu} \\\And
  Francesco Ortu \\
  University of Trieste  \\
  \texttt{francesco.ortu@phd.units.it} \\\AND
  Roya Ensafi \\
  University of Michigan \\
  \texttt{ensafi@umich.edu} \\\And
  Zhijing Jin \\
  MPI \& University of Toronto \\
  \texttt{zjin@cs.toronto.edu} \\\And
  Rada Mihalcea \\
  University of Michigan \\
  \texttt{mihalcea@umich.edu} \\
}

\begin{document}
\maketitle
\begin{abstract}
The ability of Natural Language Processing (NLP) methods to categorize text into multiple classes has motivated their use in online content moderation tasks, such as hate speech and fake news detection. However, there is limited understanding of how or why these methods make such decisions, or why certain content is moderated in the first place. 
% Due to their capacity to understand contextual information, Large Language Models (LLMs)  are well-suited for research in this area. 
To investigate the hidden mechanisms behind content moderation, we explore multiple directions: 1) training classifiers to reverse-engineer content moderation decisions across countries; 2) explaining content moderation decisions by analyzing Shapley values and LLM-guided explanations. Our primary focus is on content moderation decisions made across countries, using pre-existing corpora sampled from the Twitter Stream Grab.
% \footnote{Twitter is now officially known as X.}.
Our experiments reveal interesting patterns in censored posts, both across countries %—where censorship aligns with specific national regulations—
and over time.
%, with censorship fluctuating based on social events.
Through human evaluations of LLM-generated explanations across three LLMs, we assess the effectiveness of using LLMs in content moderation. Finally, we discuss potential future directions, as well as the limitations and ethical considerations of this work.\footnote{Our code and data 
\ifarxiv
    are at \url{https://github.com/causalNLP/censorship}.
\else
    have been uploaded to the submission system, and will be open-sourced upon acceptance.
\fi
}

\color{red}{\noindent\textbf{Disclaimer:} This paper contains examples that may be considered offensive or hateful towards certain groups.}
\end{abstract}

% I suggest reorganizing the paper to keep the "uncovering" and "understanding" parts together.
% Sec 2 Rel Work
% Sec 3 Data (current sec 4)
% Sec 4 Uncovering Censorship. Move under this section the current 3.1 and 3.2, 5.1, 6.1, 6.2. % "Reproducing censorship decisions"
%Sec 5 Understanding Censorship. Move under this section the current 3.3, 5.2, 6.3.
% Sec 6 Lessons Learned. Mover under this sec 6.4

\section{Introduction}

\begin{table}[t]
\centering
\resizebox{\linewidth}{!}{ % Ensures it fits within the column width
\normalsize
\begin{tabular}{p{4em} p{13em} p{4em} } % Adjust column widths for better text flow
\toprule
\textbf{Country} & \textbf{Censored Post} & \textbf{Type} \\ \toprule
\textbf{Germany}  & \hl{London Tube Bomber} was Taken in as a Refugee at age 15.
\hl{They are NOT Refugees they are invaders Soldiers 4 Allah}  & Harmful Censorship \\ \midrule
\textbf{France} & Give Me Your Thoughts. \hl{Military friend} believes this videos proves there were at \hl{LEAST 2 shooters} on 32nd floor  & Military Censorship \\ \midrule
\textbf{India}  & \hl{\#KashmirUnderThreat}. Use of \hl{cluster bombs} targeting civilian population at ceasefire line has \hl{exposed India \#KashmirBleeds}  & Political Censorship \\ \midrule
\textbf{Turkey}  & He is Kurdish \hl{became Christian}. That alone is reason enough 4 \hl{Iranian Mullahs} 2 hang him. These idiots are same as \hl{daesh} only shia form  & Religious Censorship \\ \midrule
\textbf{Russia} & Watch \hl{bet on football} whenever you like it. Try \hl{\#SBOBETs Virtual football} today bring the action to a new level  & Corporate Censorship \\ \bottomrule
\end{tabular}
}
\caption{Example censored tweets across five countries. This table presents verbatim examples of social media posts from five countries in our study, illustrating diverse censorship rationales. The highlighted segments denote key textual elements that are most salient for censorship classification.}

\label{tab:dataset-examples}
\vspace{-1em}
\end{table}

In an era where billions of users engage with digital platforms daily, content moderation serves as a critical governance mechanism that balances freedom of expression with the need to mitigate harm and maintain community standards \citep{grimmelmann2015virtues}. Platforms enforce moderation policies to address a wide array of issues, including terrorism, graphic violence, hate speech, explicit content, child exploitation, and fraudulent activities such as spam and fake accounts \citep{Boyer_2003, gorwa2020algorithmic, arora2023detecting}.

However, as mega-platforms scale, concerns persist regarding the transparency and accountability of moderation practices \citep{suzor2019we}. The challenges manifest at two levels: (1) the \textbf{opacity} of moderation decisions, often hidden within proprietary algorithms, platform policies, and government interventions \citep{russia-cens, Akgul2015Internet}, and (2) the \textbf{methodological limitations} of NLP-based moderation systems, which predominantly rely on surface-level keyword detection without distinguishing between \textit{mention} and \textit{usage} \citep{gligoric-etal-2024-nlp}. Moreover, the lack of interpretability of black-box models makes it difficult to verify whether content moderation decisions are justified or erroneous. While prior work has focused on improving moderation techniques, little attention has been paid to systematically analyzing the reasoning behind these decisions \citep{Kolla2024LLMModCL, huang2024content}.

To address these gaps, this work pioneers a novel investigation: deploying LLMs 
% as \textit{investigative journalists} 
to audit and interpret cross-national moderation patterns, like automatic investigative journalists. We conduct a case study on Twitter posts censored in five countries—\textit{Germany, Turkey, India, France, and Russia}—spanning from 2011 to 2020. 

We explore two research questions:
\vskip 0.05in
%\begin{enumerate}[noitemsep, left=0pt, topsep=0pt, label={\bf RQ\arabic*:}]
\noindent {\bf RQ1: Can LLMs reverse-engineer moderation decisions across different countries?}
We model censorship behavior using neural classifiers of varying architectures and scales, leveraging the content moderation dataset from \citet{censorship-dataset-aaai21} to predict whether a post was censored and to categorize it into one of six predefined censorship categories.
\vskip 0.05in

\noindent {\bf RQ2: Can explainable AI techniques applied to LLMs reveal the underlying mechanisms of content moderation decisions over time and across events?}
We employ Shapley values \citep{lundberg-shap2017} to extract the most influential entities in censored posts and use LLMs to generate interpretable explanations for the content moderation decisions \citep{bills2023language}.
%\end{enumerate}

Our findings reveal that LLM-based classifiers can effectively replicate real-world content moderation decisions, with notable cross-country variations, particularly in the distribution of misclassified censorship categories. Additionally, both Shapley values and LLM explanations help in inferring the censorship patterns, with their validity and explainability verified through human evaluation. A temporal analysis also suggests that censorship patterns align with major social and political events. Finally, we discuss recent emergent censorship behavior in reasoning models such as DeepSeek R1 and concerns about applying LLMs to real-world content moderation.

To summarize, the contributions of this work are threefold:
\begin{enumerate}[noitemsep, left=0pt, topsep=0pt]
\item We systematically reverse-engineer online content moderation decisions in five countries over a decade using LLMs, uncovering patterns in how content moderation varies across geopolitical contexts.
\item We enhance explainability of moderation decisions by applying Shapley values and LLM-guided reasoning, providing insights into the factors influencing content moderation and validating their effectiveness through human evaluation.
\item We discuss the broader implications and potential applications of this study for online content moderation.
\end{enumerate}

Our work is among the first to explain content moderation decisions across countries and time, addressing the challenges of automated content moderation and exploring the potential of LLMs in this domain.

\section{Related Work}
\label{sec:rel-work}

\paragraph{Online Moderation}  

Content moderation is crucial for maintaining online discourse, encompassing hate speech detection \cite{waseem-hovy-2016-hateful, waseem-etal-2017-understanding, founta-hatespeech}, political censorship \cite{censorship-dataset-aaai21}, and factual verification \cite{thorne-etal-2018-fever}. While investigative journalists rely on nuanced decision-making, automated moderation systems predominantly use keyword-based filtering \cite{MacKinnon_2009, censorshipinthewild-imc14}, which often lacks contextual depth. Large datasets \cite{censorship-dataset-aaai21, knockel-etal-2018-effect} provide benchmarks, but moderation policies remain fluid.

Automated systems frequently misclassify critical content, such as calls to action in social movements \cite{rogers-etal-2019-calls}, while users devise creative linguistic adaptations to bypass censorship \cite{ji-knight-2018-creative}. Traditional language models (LMs) also struggle to distinguish between the use and mention of keywords, sometimes incorrectly censoring counterspeech \cite{gligoric-etal-2024-nlp}. 

With the rise of LLMs, research has examined their effectiveness in moderation. \citet{huang2024contentmoderationllmaccuracy} evaluate LLMs on broader metrics beyond accuracy, while \citet{watchyourlanguage-icwsm24} report that LLMs outperform traditional classifiers in toxicity detection. However, moderation remains subjective, as \citet{masud2024hatepersonifiedinvestigatingrole} show that LLMs' persona-based attributes influence hate speech annotation.

\paragraph{Explainability}  

Explainability plays a crucial role in AI decision-making, particularly in content moderation and healthcare \citep{pate2023investigating}. Techniques such as causal interventions analyze neuron behaviors \cite{vigetal2020-cma, dalvi2019-whatisonegrain}, while intermediate representation translation reveals concept evolution across layers \cite{nostalgebraist2020, belrose2023eliciting}. Influence functions, LIME, and SHAP provide model explanations without modifying architectures \cite{pmlr-v70-koh17a, lundberg-shap2017}, and \citet{bills2023language} propose an explainer model for interpreting LM decisions. Activation Patching enables self-interpretation of LM representations \cite{meng2022-rome, ghandeharioun2024patchscopes}, while \citet{swayamdipta-etal-2020-dataset} examine training dynamics to assess dataset impact.

In content moderation, \citet{mathew2021hatexplain} introduced HateXplain, the first benchmark for explainable hate speech detection. However, prior work does not address censorship explainability across multiple countries and time periods. Our study takes a novel approach by treating LLMs as investigative journalists to analyze content moderation decisions across five countries. Unlike HateXplain, our work (1) focuses on content moderation, which lacks rationale-specific datasets, and (2) generates full-length explanations for censored posts. To our knowledge, this is the first large-scale study using LLMs for content moderation pattern analysis and explanation.

% , which we believe is one of the biggest contributions we bring to the field of NLP.

% \citet{nostalgebraist2020} introduced LogitLens, which is a pretty-straightforward method to interpret the intermediate representations of GPT-2 by directly translating the layerwise outputs to the vocabulary space. \citet{belrose2023eliciting} highlighted some problems in LogitLens and introduced the TunedLens which instead learns an adapter for each layer using a distallation loss over the final layer outputs.

\begin{table}[ht]
\resizebox{\linewidth}{!}{
\begin{tabular}{lccccc}
\toprule
\textbf{Feature} & \textbf{Germany} & \textbf{France} & \textbf{India} & \textbf{Turkey} & \textbf{Russia} \\ \toprule
% \# Posts & 206,637 & 25,850 & 16,280 \\
% \# Words/Post & 20.03 & 19.94 & 20.01 \\
% Post Length & 120.1 & 119.6 & 120.1 \\
% Type-Token Ratio & 0.08 & 0.16 & 0.19 \\
% Label Distribution & 1.78:1 & 1.78:1 & 1.71:1 \\
% Country Distribution & 16:15:2:2:15 & 16:15:2:2:15 & 16:15:2:2:16 \\
\# P & 39447 & 36197 & 5028 & 37243 & 4669 \\
\# W/P & 21.86 & 22.18 & 23.87 & 18.79 & 21.98 \\
Avg. Len & 131.07 & 133.58 & 142.44 & 113.21 & 131.2 \\
TTR & 0.03 & 0.03 & 0.11 & 0.05 & 0.14 \\
\% / all & 16\% & 14\% & 2\% & 15\% & 1.8\% \\ \bottomrule
\end{tabular}
}
% \caption{Detailed statistics (Num. posts, Num. words/post, Avg. Length, Type-to-token ratio, \% of all posts) about the dataset splits used in our experiments.}
\caption{Dataset Statistics. Summary of key statistics for the five countries in our dataset, including the number of posts, average words per post, average length, type-to-token ratio, and percentage of total posts.}

\label{tab:dataset-statistics}
\vspace{-1em}
\end{table}

\section{
% Multi-Country Content Moderation 
Dataset Description}
\label{sec:data}

\subsection{Data Description}

% [Background motivation of this dataset. Introduce that we are the first to bring it into NLP]

% \paragraph{Motivation}
% Due to the lack of transparency behind censorship and the lack of political literacy in many countries, we wish to introduce the concept of explainability and transparency through our work of inferring the causation behind censorship in 5 countries.
We focus on data from five countries: Germany, France, India, Turkey, and Russia. These countries were selected for their diverse national contexts and because they have the highest number of samples as shown in \citet{censorship-dataset-aaai21}, from which we can borrow our data guidelines for streaming the Twitter API.

\paragraph{Dataset Characterization.} This dataset was an artifact of the Twitter Stream Grab, which is a 1\% subsample of all the tweets (and retweets) between 2011 and 2020 July. As reported by the authors, the number of censored posts was reduced significantly from the time of their own analysis as of December 2020. We used Twitter's API to crawl the tweets\footnote{We could not use the dataset released by \citet{censorship-dataset-aaai21} because they only released the Twitter IDs and not the actual post content, which is crucial for our analysis.} much later than the publishing of the previous work. Due to this, there was a subset of tweets missed due to problems we could not control: users deleting their tweets, tweets being blocked by the Twitter API, etc. The dataset also contains multiple attributes about the posts such as which countries the tweet is censored in,  if the poster's profile was censored, and its replies or retweets. The language of all samples in the dataset is English. We create a subset of the existing overall combined test set after sampling 500 samples per country,
% \footnote{There are only 375 total samples for India, and 400 for Russia. However, for the rest of the countries, there are around 1.4k samples in total.}
exclusive of the train set. We choose at most 500 samples in order to make sure our country-wise classifications are more or less fair, as the number of censored samples is inconsistent across all 5 chosen countries. We also report our findings from the predictions over these country-wise test sets. There is no country-wise separation during the training. Note that the country-wise sets may have some overlap because a post can be censored in more than one countries. Table~\ref{tab:dataset-examples} shows some excerpts taken verbatim from the dataset along with their country of moderation and the explanations generated by GPT-4o-mini.

\begin{figure*}[t!]
\centering
    \begin{subfigure}[c]{0.45\textwidth}
        \centering
        \includegraphics[width=\textwidth]{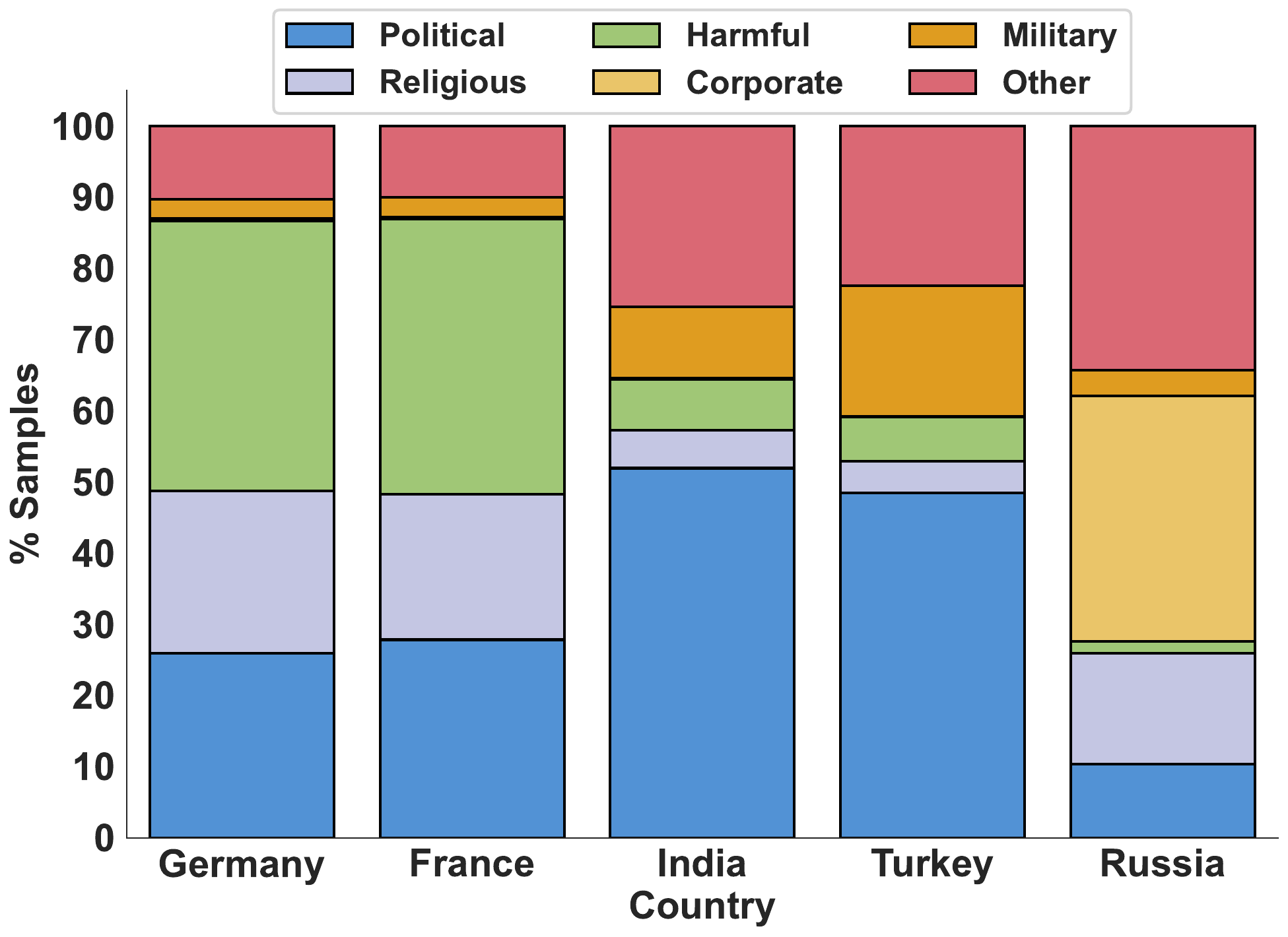}
        \caption{Category Distribution.}
        \label{fig:category_distribution}
    \end{subfigure}
    ~ \hspace{2em}
    \begin{subfigure}[c]{0.45\textwidth}
        \centering
        \includegraphics[width=\textwidth]{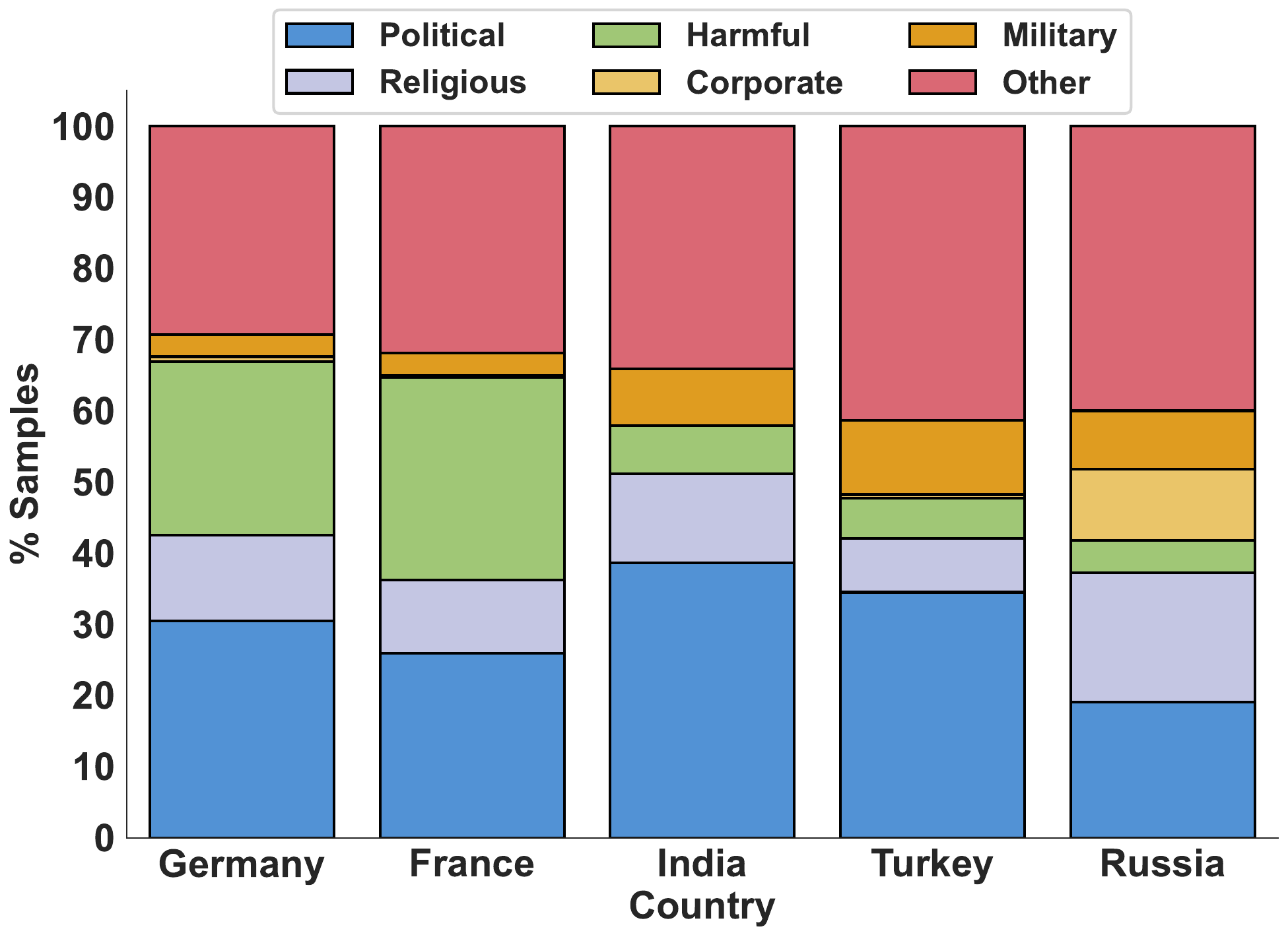}
        \caption{Misclassification Distribution. \% of misclassified samples per country: Germany-6\%, France-4\%, India-14\%, Turkey-8\%, Russia-14\%.
        %\jiarui{Instead of figures, could just be a summary of the results. Since the left figure feels like data statistics while this one feels like error analysis.}
        }
        \label{fig:error-category_dist}
    \end{subfigure}
% \caption{\textbf{Comparison between predicted categories and misclassifications.\jiarui{Rewrite the captions for Figure 1 to make it more understandable for having (a) and (b) side by side}} Fig. (a) is a \jiarui{remove the type of plots e.g., stacked bar plot} that shows the distribution of \% of samples that belong to each category index. Fig. (b) is a stacked bar plot that shows the distribution of \% misclassified samples that belong to previously defined categories.}
\caption{Category Distribution and Misclassification Patterns Across Countries.  
Figure (a) illustrates how samples are distributed across different censorship categories, providing insights into the relative prevalence of each category within the dataset. 
Figure (b) depicts the classified categories for our best-performing model across countries, showing how misclassified samples are distributed among the predefined categories. Notably, compared with the ground truth (Figure (a)),  a substantial portion of the misclassified instances fall into the \emph{Other} category, suggesting that these cases might be harder to categorize accurately, possibly due to overlapping features.}
% \francesco{Clarify whether (b) represents the average classification across all models or results from a specific model. Add this info in the caption.} }

\vspace{-1em}
\end{figure*}

\subsection{Data Statistics}

\paragraph{Overview}
In Table~\ref{tab:dataset-statistics}, we present general statistics of the dataset introduced earlier. The type-token ratio (TTR) is computed as \texttt{\#unique words / total \#words}. A lower TTR suggests reduced token diversity, potentially indicating repeated entities that are significant for censorship across multiple posts. Approximately $16\%$ of the samples were censored in Germany, $14\%$ in France, $2\%$ in India, $15\%$ in Turkey and , $2\%$ in Russia. India and Russia have relatively fewer posts than the other three countries. The validation sets are about $12\%$ of the training set size, while the test sets comprise approximately $8\%$.

\paragraph{Topic Categories}

According to the taxonomy provided by the RSF methodology for \href{https://rsf.org/en/methodology-used-compiling-world-press-freedom-index-2024?year=2024&data_type=general}{calculating the Press Freedom Index} and \citet{Warf2010}, we categorize censorship into six key types that account for a significant portion of censored posts:

\begin{enumerate}[noitemsep, nolistsep, topsep=0pt]
\item Political censorship
\item Religious censorship (including cultural restrictions and religion-based constraints)
\item Harmful content (racism, hate speech, obscenity, etc.)
\item Corporate censorship (including IP laws, gambling, spam)
\item Military censorship (content deemed harmful to national security or containing confidential military intelligence)
\item Other (to account for outliers that do not fit into the primary categories -- relatively random, spam and other topics that LLMs cannot associate with the other 5)
\end{enumerate}

To classify censored posts from our target countries $\mathcal{C}$, we use GPT-4o-mini, leveraging its strong world knowledge and ability to model distributional patterns across different nations. We manually evaluate its predictions on a random sample of 50 examples per language and find its classification performance to be nearly perfect.

Figure~\ref{fig:category_distribution} illustrates the predicted categorizations\footnote{Refer to Appendix~\ref{appendix:category_examples} for details on common keywords associated with each category and the rationale behind their selection.}. Interestingly, Germany and France exhibit similar distributions, with the highest proportion of Category 3 censorship. Russia has the highest levels of Category 4 and 6 censorship, while India and Turkey display similar patterns, with the most censorship occurring under Category 1. 

All analyses related to these categories are performed on the validation set, as it contains a relatively higher number of samples than the test set, particularly for countries where data is already scarce.

% \jiarui{task 1: reverse content moderation decisions; }

% \jiarui{task 2: revealing/interpret hidden mechanisms: censorship category differences across countries (mention this in the dataset description, and we echo it in experiment result section); shapley value more real less readable; LLm explanation generation more readable less faithful; human evaluation}

\section{Task Formulation}
\label{sec:task-form}
In this section, we formulate two key tasks based on our research questions: (1) reverse-engineering content moderation decisions and (2) leveraging LLMs to uncover hidden patterns behind these decisions.

% \textcolor{red}{echo with "uncovering" and "understanding" by the task formulation. Then also reflect this in the Intro}

\subsection{Task 1: Reverse-Engineering Content Moderation}
% \jiarui{Germany, France, India, Turkey, Russia, increasing political censorship}
% \subsection{Uncovering Censorship}
% \paragraph{Task 1: Detecting Censorship}
\label{subsec:task1}
To achieve our goal of understanding censorship, we start with predicting if a post was censored in a country $c \in \mathcal{C}$. $\mathcal{C}$ refers to the set of test sets $\{$\texttt{Germany, France, India, Turkey, Russia}$\}$. Within our dataset, for each censored post, there can be multiple labels for the country. For example, a post can be censored in both Germany and France.

We train a multi-label classifier using five encoder-based language models (LMs): BERT-Tiny \cite{bhargava2021generalization}, BERT Base and m-BERT \cite{devin18-bert}, XLM-RoBERTa \cite{conneau-etal-2020-xlmr}, and RoBERTa \cite{roberta-2019}. Additionally, we include two small-scale decoder-based LLMs, Pythia (1B) \cite{biderman2023pythia} and Llama 3.2 (1B), for censorship prediction experiments. These small-scale models were chosen for their memory efficiency, making them well-suited for resource-constrained applications. We report the Weighted F1 metric for this task.

Beyond trained models, we also evaluate larger state-of-the-art LLMs in a zero-shot setting without training on the dataset, including Aya-23-8B \citep{aryabumi2024aya23openweight}, Llama-3.1-8B-Instruct \citep{dubey2024llama3herdmodels}, GPT-4o-mini, and GPT-4o \citep{gpt4o-openai}. Table~\ref{tab:model-list} lists all LLMs used for prompting, along with their knowledge cutoffs. We conduct classification experiments to assess the zero-shot censorship detection performance of these models. Additional experimental details can be found in Appendix~\ref{appendix:exp_details}.

\begin{table*}[t!]
\resizebox{\linewidth}{!}{
\centering
\begin{tabular}{l l c c c c c c c}
\toprule
\multirow{2.5}{*}{\textbf{Country}} & \multirow{2.5}{*}{\textbf{Split}} 
  & \multicolumn{5}{c}{\textbf{\textit{Encoder Language Models}}} 
  & \multicolumn{2}{c}{\textbf{\textit{Decoder Language Models}}} \\ % & \multicolumn{4}{c}{\textbf{\textit{Large Language Models}}} \\
\cmidrule(lr){3-7}\cmidrule(lr){8-9}
& & \textbf{BERT-Tiny} & \textbf{BERT Base} & \textbf{XLM-R} 
  & \textbf{m-BERT} & \textbf{RoBERTa Base} 
  & \textbf{Pythia} & \textbf{Llama 3.2} \\ % & \textbf{Aya-23} & \textbf{Llama-3.1} & \textbf{GPT-4o-mini} & \textbf{GPT-4o} \\

\midrule
% \midrule
\multirow{2}{*}{\textbf{Germany}}
  & \textbf{Val.} & 76.91 & 89.89 & 88.21 & 89.39 & 90.93 & 93.81 & 92.76 \\
  & \textbf{Test} & 78.19 & 91.26 & 89.27 & 90.03 & 91.64 & 94.53 & 93.69 \\
\midrule
\multirow{2}{*}{\textbf{France}}
  & \textbf{Val.} & 76.90 & 90.62 & 88.88 & 90.21 & 92.03 & 94.86 & 93.94 \\
  & \textbf{Test} & 78.35 & 91.73 & 89.87 & 90.77 & 92.67 & 95.51 & 95.19 \\
\midrule
\multirow{2}{*}{\textbf{India}}
  & \textbf{\textbf{Val.}} &  0.41 & 79.87 & 75.22 & 79.28 & 80.86 & 87.11 & 78.34 \\
  & \textbf{Test} &  0.57 & 78.83 & 72.56 & 77.19 & 77.95 & 86.64 & 85.97 \\
 \midrule
\multirow{2}{*}{\textbf{Turkey}}
  & \textbf{Val.} & 73.05 & 89.20 & 86.36 & 88.25 & 88.92 & 92.05 & 91.15 \\
  & \textbf{Test} & 73.55 & 89.25 & 87.15 & 88.48 & 89.77 & 92.04 & 91.82 \\
\midrule
\multirow{2}{*}{\textbf{Russia}}
  & \textbf{Val.} &  0.00 & 79.73 & 73.08 & 77.13 & 79.11 & 85.59 & 82.08 \\
  & \textbf{Test} &  0.00 & 77.19 & 73.33 & 78.26 & 79.93 & 85.50 & 85.61 \\
\midrule
 \midrule
 \addlinespace
\multirow{2}{*}{\textbf{Aggr.}}
  & \textbf{Val.} & 69.77 & 89.11 & 86.75 & 88.43 & 89.79 & 91.64 & 93.00 \\
  & \textbf{Test} & 70.71 & 89.74 & 87.51 & 88.80 & 90.36 & 92.94 & 93.39 \\
\bottomrule
\end{tabular}

% \begin{tabular}{lcc|cccccccccc}
% \toprule
% \multirow{2}{*}{\textbf{Model}} & \multicolumn{2}{c}{\textbf{Aggr.}} & \multicolumn{2}{c}{\textbf{Turkey}} & \multicolumn{2}{c}{\textbf{Russia}} & \multicolumn{2}{c}{\textbf{Germany}} & \multicolumn{2}{c}{\textbf{France}} & \multicolumn{2}{c}{\textbf{India}} \\ \cmidrule{2-13}
% & \textbf{Val.} & \textbf{Test}& \textbf{Val.} & \textbf{Test}& \textbf{Val.} & \textbf{Test}& \textbf{Val.} & \textbf{Test}& \textbf{Val.} & \textbf{Test} & \textbf{Val.} & \textbf{Test} \\ \midrule
% \multicolumn{3}{l}{\textbf{\textit{Smaller Language Models}}} \\
% \quad BERT-Tiny & 69.77 & 70.71 & 73.05 & 73.55 & 0 & 0 & 76.91 & 78.19 & 76.9 & 78.35 & 0.41 & 0.57 \\
% \quad BERT Base & 89.11 & 89.74 & 89.2 & 89.25 & 79.73 & 77.19 & 89.89 & 91.26 & 90.62 & 91.73 & 79.87 & 78.83 \\
% \quad XLM-RoBERTa & 86.75 & 87.51 & 86.36 & 87.15 & 73.08 & 73.33 & 88.21 & 89.27 & 88.88 & 89.87 & 75.22 & 72.56 \\
% \quad m-BERT & 88.43 & 88.8 & 88.25 & 88.48 & 77.13 & 78.26 & 89.39 & 90.03 & 90.21 & 90.77 & 79.28 & 77.19 \\
% \quad RoBERTa Base & 89.79 & 90.36 & 88.92 & 89.77 & 79.11 & 79.93 & 90.93 & 91.64 & 92.03 & 92.67 & 80.86 & 77.95 \\ \hdashline
% \multicolumn{3}{l}{\textbf{\textit{Larger Language Models}}} \\
% \quad \ul{Pythia (1B)} & \ul{91.64} & \ul{92.94} & 92.05 & 92.04 & 85.59 & 85.5 & 93.81 & 94.53 & 94.86 & 95.51 & 87.11 & 86.64 \\
% \quad \textbf{Llama 3.2 (1B)} & \textbf{93} & \textbf{93.39} & 91.15 & 91.82 & 82.08 & 85.61 & 92.76 & 93.69 & 93.94 & 95.19 & 78.34 & 85.97 \\ \bottomrule
% \end{tabular}
}
\caption{Country-wise Accuracies.  
This table presents the accuracy scores for each model across different countries, evaluated using weighted F1 scores on the test set. "Aggr." refers to the results on the aggregated test set, which serves as the primary criterion for determining the overall best-performing model.
}
\label{tab:countrywise-acc}
\vspace{-1em}
\end{table*}

\begin{table}[t]
\small
\resizebox{\linewidth}{!}{
    \centering
    % \begin{tabular}{lcccccc}
    % \toprule
    %      \textbf{LLM} & \textbf{Aggr.} & \textbf{Germany} & \textbf{France} & \textbf{India} & \textbf{Turkey} & \textbf{Russia} \\ \midrule
    %      Aya & 47.6 & 37 & 49 & 56 & 62 & 34 \\
    %      Llama & 18.2 & 11 & 7 & 29 & 36 & 8 \\
    %      GPT-4o-mini & 53.6 & 49 & 42 & 66 & 77 & 34 \\
    %      GPT-4o & 53.4 & 41 & 36 & 66 & 80 & 44 \\ \bottomrule
    % \end{tabular}
    \begin{tabular}{lcccc}
    \toprule
         \textbf{Country} & \textbf{Aya-23} & \textbf{Llama-3.1} & \textbf{GPT-4o-mini} & \textbf{GPT-4o} \\ \midrule
         \textbf{Germany} & 37.0 & 11.0 & 49.0 & 41.0 \\
         \textbf{France}  & 49.0 & 7.0  & 42.0 & 36.0 \\
         \textbf{India}   & 56.0 & 29.0 & 66.0 & 66.0 \\
         \textbf{Turkey}  & 62.0 & 36.0 & 77.0 & 80.0 \\
         \textbf{Russia}  & 34.0 & 8.0  & 34.0 & 44.0 \\
         \midrule
         \midrule
        \addlinespace
         \textbf{Aggr.} & 47.6 & 18.2 & 53.6 & 53.4 \\ 
         \bottomrule
    \end{tabular}
}
\caption{Country-wise accuracy of additional decoder language models on the validation set. This table reports accuracy scores across different countries, with results constrained by computational limitations. For GPT-* models, predictions are based on a subset of 500 samples per country due to the high cost of API queries.}

    \label{tab:llm-acc}
\vspace{-1em}
\end{table}

\subsection{Task 2: Explaining Content Moderation}
\label{subsec:task2}
Beyond analyzing predictions and data distribution, we also aim to explain the reasoning behind these decisions. In this task, we leverage the predictions from Task 1 to compute Shapley values, aiming to identify patterns across countries and over time. Additionally, we employ LLMs to generate explanations for model censorship predictions and manually verify their generations.

\paragraph{Task 2.1: Shapley Explanations.}  
Building on prior work in explainability, we apply SHAP \cite{lundberg-shap2017} to identify the most globally relevant entities for each country by computing Shapley values for token contributions to predictions. To align this with our research motivation, we conduct this analysis separately for each country's subset. We then compare these findings with the year-wise distribution of censorship data and contextualize them within real-world social events.

\paragraph{Task 2.2: LLM-Guided Explanations.}  
With the rise of LLMs and their ability to model world knowledge efficiently, we simulate a content-moderation setting where LLMs provide reasoning for why a post should be moderated or censored online. Building on recent LLM interpretability research \cite{bills2023language}, we employ similar techniques to extrapolate relevant entities into concise yet ``approximate'' explanations %\footnote{We refer to these explanations as ``approximate'' due to the inherently generic nature of LLM outputs in domain-specific contexts \cite{kunz-kuhlmann-2024-properties, lee-etal-2024-towards, yadav-etal-2024-tox}.} 
for model predictions \cite{kunz-kuhlmann-2024-properties, lee-etal-2024-towards, yadav-etal-2024-tox}. These explanations offer a high-level understanding of the topic and help evaluate whether the LLMs listed in Table~\ref{tab:model-list} can contribute to content moderation. To generate possible rationales for censorship, we prompt three LLMs to explain why a given post should be moderated. We then conduct a human evaluation to assess the effectiveness of these explanations across all countries.

\section{Results and Findings}
\label{sec:results}

\vspace{-0.4em}
\subsection{Results for Reproducing Content Moderation Decisions}

\paragraph{How Accurately Can LM Classifiers Reproduce Real-World Content Moderation Decisions?}

As discussed in \S\ref{subsec:task1}, we present our results for censorship prediction in Tables~\ref{tab:countrywise-acc} and \ref{tab:llm-acc}. Table~\ref{tab:countrywise-acc} shows that even encoder-based LMs achieve high F1 scores, with RoBERTa leading at 90.36 points, followed by Llama 3.2 at 92.94 points, and Pythia 1B achieving the highest score of 93.39 points. We later use Pythia 1B, the best-performing model, as the anchor model for generating Shapley values for each token in a given text, to help identify entities that significantly influence model predictions. 
Table~\ref{tab:llm-acc} demonstrates that, without training, model performance remains low across all countries. Content moderation in Germany, France, and Russia is more challenging to classify compared to India and Turkey. Among the four models tested, GPT-4o and GPT-4o-mini achieve the highest classification performance, highlighting the non-trivial nature of the task.

% \begin{figure}[h]
% \resizebox{\linewidth}{!}{
%     \centering
%     \includegraphics[width=\linewidth]{assets/error-category_distribution.pdf}
% }
%     \caption{\textbf{Misclassification Distribution across Countries.} This stacked bar plot shows the distribution of the \% of misclassified samples that belong to the previously defined categories.}
%     \label{fig:error-category_dist}
% \vspace{-2em}
% \end{figure}

\paragraph{Do Censorship Patterns Vary across Countries?} To dive deeper into the patterns in censorship on a geopolitical scale, we take a look at Table~\ref{tab:countrywise-acc} that shows our best-performing model's accuracy on samples censored in specific countries only. We find that this is very dependent on the number of samples and hence does not show any strong patterns (we discuss some country-specific social events that can affect censorship in \S~\ref{subsec:shap-analysis}.
%\jiarui{Any country-specific events? Keywords? Hashtags? I feel like we should not have this paragraph in the main text if there are no interesting findings.} 
However, in general it was easiest to predict censorship in a French context in comparison to other contexts while being very similar to German contexts (geopolitically speaking). We show a t-SNE visualization of all censored posts in our training set differentiated per country in Fig.~\ref{fig:tsne-viz}

\paragraph{Where do LMs tend to fail?}
% \jiarui{Incomplete}
% \jiarui{Neemesh: Could you mention the misclassification ratio overall for each language in Figure 1.b here?}
To understand if there are any general patterns in the examples where models frequently misclassify, we look at Figure~\ref{fig:error-category_dist} that shows the distribution of misclassified samples per category for each country. Generally speaking, the models had a hard time predicting text of Type 6 (Category: Other) Censorship across all countries, which makes sense due to the seemingly ambiguous nature of those posts (see: App.~\ref{appendix:category_examples} for more information on the `Other' category).
% \jiarui{Neemesh, could you provide some example subtopics for "Other" here? For example, sensitive contents, etc.?}
This hints towards the open-ended nature of that category which introduces some difficulty. Surprisingly, the Indian and Turkish sets had a majority of Type 1 Censorship while having the highest amount of samples in that category. 
% \jiarui{I still think this paragraph should include more concrete analyses. Moved my previous comments here: 1. Provide an analysis of commonly misclassified and correctly classified topics for each censorship category.
% 2. Clarify what is included under the "other" category of censorship.
% 3. Highlight that political censorship also has a high misclassification rate and analyze the reasons behind it. For instance, if it is described as "surprising," explain why.
% 4. Cite relevant papers discussing hard-to-recognize tweets that should be censored.}

\setlength{\abovecaptionskip}{4pt}
\begin{table*}[t]
 % \resizebox{\textwidth}{!}{
 
    \centering \small
    \begin{tabular}{lccccccc}
    \toprule
         \multirow{2}{*}{\textbf{Country}} & \multirow{2}{*}{\textbf{Pref. Order}} & \multicolumn{3}{c}{\textit{\textbf{Fluency}}} & \multicolumn{3}{c}{\textit{\textbf{Helpfulness}}} \\ \cmidrule(lr){3-5}  \cmidrule(lr){6-8}
         & & \textbf{Aya-23} & \textbf{Llama-3.1} & \textbf{GPT-4o-mini} & \textbf{Aya-23} & \textbf{Llama-3.1} & \textbf{GPT-4o-mini} \\ \midrule
         
         \textbf{Germany} & 3>2>1 & \textbf{3.50} & 3.33 & 3.33 & 3.39 & \underline{\textbf{3.61}} & 3.44 \\
         \textbf{France} & 1>2>3 & \textbf{3.56} & 3.22 & 3.33 & 3.44 & \underline{\textbf{3.56}} & 3.44 \\ 
         \textbf{India} & 1=2>3 & 3.78 & 3.78 & \textbf{3.89} & \underline{\textbf{3.94}} & 3.89 & 3.83 \\ 
         \textbf{Turkey} & 1>2>3 & \textbf{4.06} & 3.78 & 3.94 & \underline{\textbf{3.94}} & 3.83 & 3.89 \\ 
         \textbf{Russia} & 3>1>2 & 3.78 & 3.72 & \textbf{4.22} & 3.06 & 3.33 & \underline{\textbf{3.44}} \\ 
         \midrule
         \midrule
         \addlinespace
         \textbf{Overall} & 1>3>2 & \textbf{3.74} & 3.57 & \textbf{3.74} & 3.55 & \underline{\textbf{3.64}} & 3.61 \\
         \bottomrule
    \end{tabular}

% \begin{tabular}{l cc cc cc cc cc cc}
% \toprule
% \multirow{2}{*}{\textbf{LLM}} 
% & \multicolumn{2}{c}{\textbf{Overall}}
% & \multicolumn{2}{c}{\textbf{Germany}} 
% & \multicolumn{2}{c}{\textbf{France}} 
% & \multicolumn{2}{c}{\textbf{India}} 
% & \multicolumn{2}{c}{\textbf{Turkey}} 
% & \multicolumn{2}{c}{\textbf{Russia}} \\

% \cmidrule(lr){2-3} \cmidrule(lr){4-5} \cmidrule(lr){6-7} \cmidrule(lr){8-9} \cmidrule(lr){10-11} \cmidrule(lr){12-13}
% & \textbf{Fluency} & \textbf{Helpfulness} 
% & \textbf{Fluency} & \textbf{Helpfulness} 
% & \textbf{Fluency} & \textbf{Helpfulness} 
% & \textbf{Fluency} & \textbf{Helpfulness} 
% & \textbf{Fluency} & \textbf{Helpfulness} 
% & \textbf{Fluency} & \textbf{Helpfulness} \\

% \midrule

% \textbf{Aya-23}     & \textbf{3.74} & 3.55 & \textbf{3.50}  & 3.39  & \textbf{3.56}  & 3.44  & 3.78  & \textbf{3.94}  & \textbf{4.06}  & \textbf{3.94}  & 3.78  & 3.06  \\
% \textbf{Llama-3.1}   & 3.57 & \textbf{3.64} & 3.33  & \textbf{3.61}  & 3.22  & \textbf{3.56}  & 3.78  & 3.89  & 3.78  & 3.83  & 3.72  & 3.33  \\
% \textbf{GPT-4o-mini} & \textbf{3.74} & 3.61 & 3.33  & 3.44  & 3.33  & 3.44  & \textbf{3.89}  & 3.83  & 3.94  & 3.89  & \textbf{4.22}  & \textbf{3.44}  \\

% \midrule
% \textbf{Pref. Order}  & \multicolumn{2}{c}{1>3>2} & \multicolumn{2}{c}{3>2>1} & \multicolumn{2}{c}{1>2>3} & \multicolumn{2}{c}{1=2>3} & \multicolumn{2}{c}{1>2>3} & \multicolumn{2}{c}{3>1>2} \\ 

% \bottomrule
% \end{tabular}

% }
    \caption{Human Evaluation of Model-Generated Explanations.  
    This table presents the results of a human evaluation conducted on the censorship explanations quality produced by three different LLMs. The evaluation was performed on 15 randomly selected samples per country, with each sample receiving two independent annotations. We made sure there was no annotator bias by masking the LLM names in the annotator form.
    % \jiarui{have model names on the table; add one sentence in the main text saying "we made sure there is no bias for annotators to see the models"}
    }
    \label{tab:human_eval}
\vspace{-1em}
\end{table*}

\subsection{Analysis of Shapley-Based Explanations}
\label{subsec:shap-analysis}

% \begin{figure}[h]
% \resizebox{\linewidth}{!}{
%     \centering
%     \includegraphics[width=\linewidth]{assets/country_accuracy.pdf}
% }
%     \caption{\textbf{Accuracy Distribution across Countries.} Accuracy of the best-performing model on our country-wise test sets distributed in a sorted order. We also show the \% samples per country in the test set on the bars.}
%     \label{fig:country-accuracy}
% \vspace{-1em}
% \end{figure}

\paragraph{How Do Shapley Values Help in Inferring Censorship Patterns?}  
Figure~\ref{fig:individual_shap_bar} presents the top 20 most influential entities identified by Shapley values for each country. These values help approximate which social events are (1) most common in the dataset and (2) have the strongest positive impact on censorship predictions. Previous studies have demonstrated the effectiveness of Shapley values in identifying key training instances \cite{ghorbani2019data, schoch-etal-2023-data}, and we adopt a similar approach for our task.

We observe a clear alignment between censored content and real-world events. For instance, the "Kavanaugh Hearings" (likely Type 1 censorship) appear frequently in Germany and France (Figures~\ref{fig:germany_bar}, \ref{fig:france_bar}) and are associated with 2018, alongside various racist and anti-religious entities. In Turkey, entities related to the controversial TV series "After Daesh" and the "Kacmaz Family" (likely Types 5 and 1, respectively) appear in the 2017 censorship landscape (Figure~\ref{fig:turkey_bar}). In India (Figure~\ref{fig:india_bar}), religion and politics are heavily interlinked \cite{pol-rel-india, inter-rel-pol} and thus censorship heavily influenced by religious topics is also political, with frequent references to "MSG"\footnote{Unlike the chemical MSG, this refers to a religious group in India that is heavily censored and scrutinized by the government and people for various reasons.}. Meanwhile, Figure~\ref{fig:russia_bar} reveals strong censorship targeting religious, political, and possibly gambling or spam-related content (that comes under Type 6), which is strongly condemned by the Russian state.
% \jiarui{Moved the previous comments here: (1) The current discussion focuses on too few entities. What about other entities? (2) Could we come up with a more straightforward visualization for the correspondence between entities and the events behind them for each country? For example, we can label the censorship category that each entity belongs to, as well as the events behind certain entities, in Figure 4. We can also provide a brief background (and possible reasons for why it was censored) for each conspicuous event in the Appendix. (3) Direct readers to further discussion in Appendix xxx.}

\begin{figure}[h]
\resizebox{\linewidth}{!}{
    \centering
    \includegraphics[width=\linewidth]{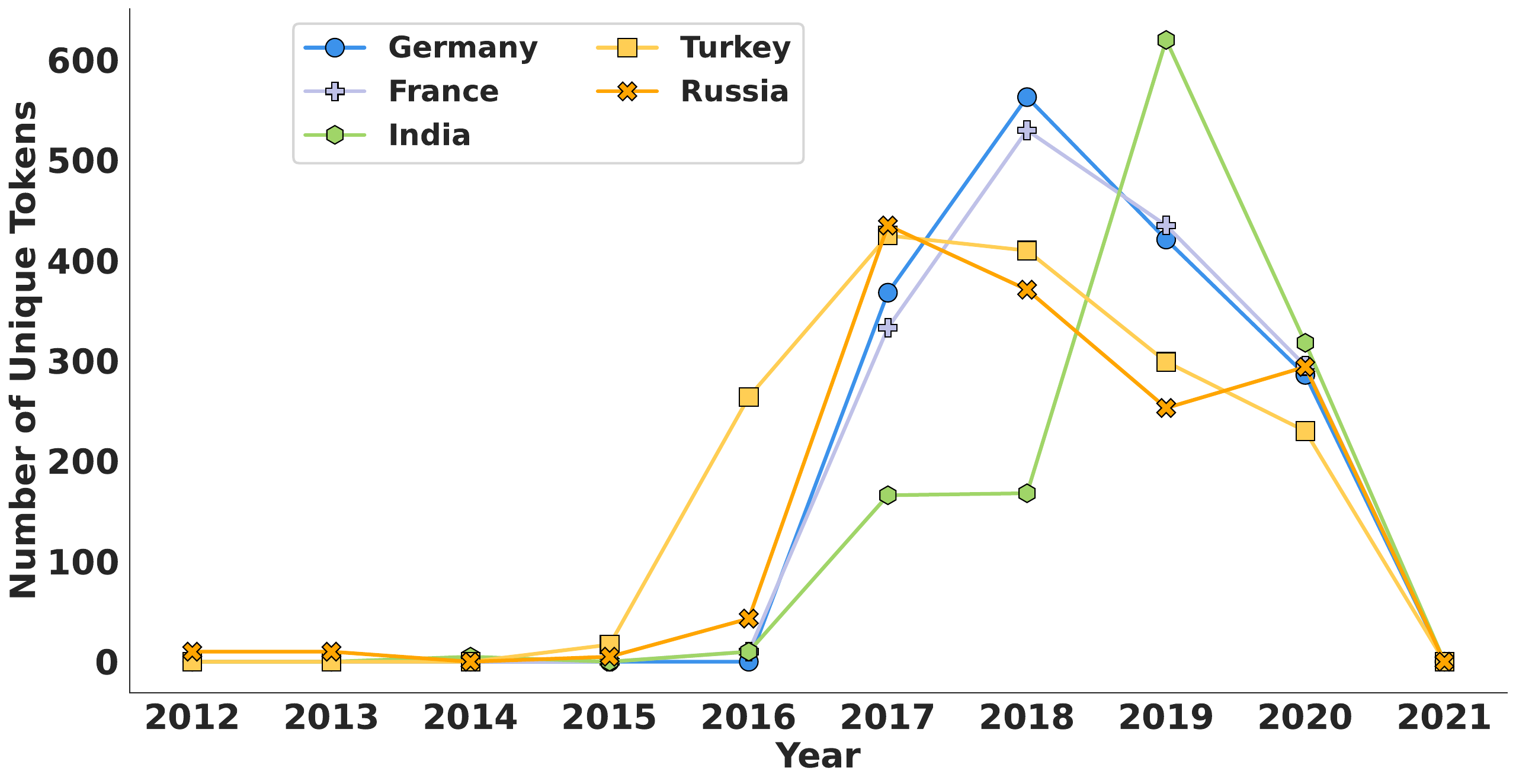}
}
\caption{Unique Token Distribution Over Time.  
The figure illustrates the number of unique tokens that played a crucial role in the model's censorship predictions across different years and countries. Notably, peaks in the distribution often coincide with political or societal events that may have caused increased censorship activity.}

    \label{fig:yearwise_country_uniquetokens}
\vspace{-1em}
\end{figure}

\paragraph{Do Censorship Patterns Vary over the years?} To address the question if LMs can model real-life social events by relating them with censorship, from simple training, we decided to look at the number of unique tokens important for censorship every year, for each country. We believe that the number of unique tokens are directly proportional to the number of censored posts -- the tokens are: 1) crucial to changing the model's prediction; and, 2) extracted directly from the posts. Simply put, the total number of unique tokens is proportional to the total number of tokens, which is the sum of the number of tokens in censored posts + non-censored posts.
% Specifically, we define $\hat{\mathbf{N}}^{y}_{c}$ to be the number of unique tokens, in year $y$ and country $c$, deemed important by looking at the top 5 Shapley weights every post. Let $\mathbf{N}^{y}_{c}$ refer to the total number censored posts in year $y$ by country $c$.
% \begin{equation}
%     \hat{\mathbf{N}}^{y}_{c} \propto \mathbf{N}^{y}_{c}
%     \label{eq:unique_tokens_proof}
% \end{equation}

% Simply put, total number of unique tokens is proportional to total number of tokens, which is the sum of number of tokens in censored posts + non-censored posts. The number of unique tokens are directly proportional to the number of censored posts -- the tokens are: 1) crucial to changing the model's prediction; and, 2) extracted directly from the posts.

Figure~\ref{fig:yearwise_country_uniquetokens} shows the distribution of $\hat{\mathbf{N}}^{y}_{C}$, where $y \in [2011-2020]$ and $c \in \mathcal{C}$. Russia and Turkey reached a peak in 2017, whereas France and Germany reached a peak in 2018, and India in 2019. This might point to certain social events that caused an influx of censorship to maintain online neutrality. Our findings from the previous question strengthen our observations here\footnote{We want to mention that although this distribution seems to contradict the category similarities for Turkey and India, it is not exactly contradictory. There may not be an overlap of social events, but there might be similar censorship laws/events taking place in the two regions at different times.}.
% \jiarui{Can we find 20 most important entities using SHAP for tweets for each year? The current discussions for this still looks shallow and not convincing.}

\subsection{Analysis of LLM-Generated Explanations}

\paragraph{How Do Topic Categories Correlate with Censorship?}
As described in \S\ref{sec:data}, we prompt GPT-4o-mini to classify censored posts into one of six censorship categories. By examining the category distributions in Figures~\ref{fig:category_distribution} and \ref{fig:error-category_dist}, we observe that while Germany and France exhibit similar category distributions, their underlying contexts differ. For instance, detecting Type 3 censorship is more challenging in France than in Germany, possibly due to differences in textual characteristics and stricter online moderation laws in Germany, such as the NetzDG law \cite{mchangama2019digital, jaki2019rightwing-germany, paasch2021insult}. A similar pattern emerges between India and Turkey, which share comparable category distributions. However, predicting Type 1 censorship appears more difficult in the Indian context than in Turkey, highlighting potential differences in regulatory frameworks and linguistic nuances.
% \jiarui{Any literature supporting this?}

\paragraph{Can LLMs Explain Content Moderation Decisions Effectively?}  
We explore the potential of using LLMs within a moderation framework to assist content moderators. Specifically, we generate post-specific explanations using the LLMs listed in Table~\ref{tab:model-list} and assess their effectiveness through human evaluation on a small subset of samples from all five countries. For this task, we conduct a small-scale human evaluation\footnote{See Appendix \S\ref{appendix:human_eval} for details on the complete human evaluation process.} on 45 samples per country (15 samples per LLM across three LLMs). Five expert evaluators participated in the evaluation, with two evaluators assigned per country across all metrics and samples. Evaluators rated the generations based on three key metrics: \textit{LLM Preference Rating}, \textit{Fluency}, and \textit{Helpfulness}. Table~\ref{tab:human_eval} presents an overview of our results, averaged across all samples and both annotations per country. The final row, indexed as \textit{Overall}, reports the average across all countries.

Interestingly, most annotators preferred the generations produced by Aya-23 (LLM 1) over GPT-4o-mini (LLM 3) and Llama-3.1 (LLM 2). This preference may be attributed to Aya-23's enhanced multilingual capabilities, which could improve its ability to process multicultural content across diverse countries. This shows that Aya-23 can be used as a decent helper for a content moderator! %, despite contrary findings by \citet{liu-etal-2024-multilingual}.

\vspace{-0.4em}
\subsection{Discussions}
\label{sec:discussion}

\paragraph{How Predictable and Explainable Is Censorship?}
% We find that although there are clear patterns across countries, they are not intuitive or feasible to infer without the help of automated systems, such as LMs and LLMs in our case.
% \textit{Without the use of LLMs in predicting near-accurate categories of censored content or generating explanations behind such censorship, it would have been unfeasible to do without a large amount of labour and cost.}

While censorship appears to be predictable (see Table~\ref{tab:countrywise-acc}), it remains largely unexplained. LMs perform well in predicting censorship decisions, yet Table~\ref{tab:llm-acc} highlights the non-trivial nature of this task, where even SOTA LLMs struggle to surpass random predictions. Table~\ref{tab:human_eval} further demonstrates that although LLMs can generate possible explanations, they are, on average, only about 72\% helpful. No single explainability method can fully uncover the underlying rationale behind censorship laws in different countries. Instead, multiple findings must be synthesized to derive meaningful observations. The apparent predictability of censorship is obscured by the contextual variability of censorship rules across countries. For instance, the same category of censorship may be easier to predict in France but more challenging in Germany due to stricter regulations requiring implicit and harder-to-detect contextual cues \cite{mchangama2019digital, jaki2019rightwing-germany, paasch2021insult}.

\paragraph{Does Censorship Follow Any Patterns?}  
Our analysis reveals no consistent censorship pattern across countries. However, Figure~\ref{fig:category_distribution} shows that certain country pairs, such as Turkey-India and Germany-France, exhibit similar category distributions. There is no clear pattern linking these pairs unless considered within a geopolitical context, suggesting a possible correlation between their censorship policies.

In contrast, censorship patterns emerge over time (see Figure~\ref{fig:yearwise_country_uniquetokens}). These trends can be attributed to both global and local events (as discussed in \S~\ref{subsec:task2}) that may conflict with the moderation guidelines of specific countries. We hypothesize that such events lead to a surge in online content, increasing the number of censored posts. For example, political content faces significantly more censorship in India and Turkey, whereas harmful content is more frequently censored in Germany and France.

\paragraph{Can LLMs Infer the Causes of Censorship?}  
We conduct multiple analyses to evaluate the ability of LLMs to infer causation behind censorship. Our observations indicate that while LLMs effectively categorize censored posts into predefined categories, they struggle to decipher entities that require deeper regional knowledge. For instance, in the Indian Shapley plot, entities related to "MSG" resemble spam-like content, making interpretation challenging (see Figure~\ref{fig:india_bar}).  

Figure~\ref{fig:wordcloud_categories} illustrates the accuracy of LLMs in categorizing censored posts. Results from human evaluation (Table~\ref{tab:human_eval}) show that, on average, LLMs are only moderately helpful, with a highest Helpfulness score of 3.64. This limitation arises from the implicit and context-specific nature of censored content, which often requires regional expertise.  

Despite these challenges, LLMs remain useful by (1) identifying underlying themes and patterns that might be overlooked by traditional analysis methods and (2) providing a rough understanding of implicit censorship rules. Additionally, they offer valuable insights for a deeper comparative analysis of censorship regulations across different countries.

\section{Opening Directions for Future Work}
Our work opens several avenues for future research, particularly in using LLMs for content moderation, explaining the decisions behind censorship, and detecting censorship across both geographical and temporal scales. Given the highly sensitive nature of this task, ensuring that LLM-generated explanations remain faithful is crucial \cite{turpin23-llm-unfaithful}. With the growing focus on interpretability, we anticipate that future research can build upon our findings using advanced techniques such as activation patching \cite{meng2022-rome} and circuit discovery \cite{conmy23-circuit_discovery}. Furthermore, expanding censorship analysis to more countries while balancing dataset distributions across regions could provide deeper insights into how multilingualism influences censorship practices.

\section{Conclusion}
This study analyzes content moderation mechanisms across multiple countries, using LLMs for classification and explainability. Our findings highlight discrepancies in moderation policies, shaped by geopolitical, temporal, and event-specific factors. Using explainability techniques like Shapley values and LLM-guided reasoning, we identify key censorship patterns, showing that while LLMs can replicate moderation practices, they struggle with nuanced, context-dependent rules. These insights emphasize the need for greater transparency in AI-driven moderation. Future research should focus on improving interpretability and ensuring the responsible deployment of LLMs in content governance.

% We perform a detailed study about investigating patterns in Censorship across multiple countries and years. We experiment with LMs as well as small-scale LLMs for predicting the censorship status of Twitter posts across multiple countries. We use these results to make inferences about the possible patterns behind Censorship rules in different countries, and how well we can predict the happening of social events with this. We make use of LLMs to classify censored posts from our test set into defined categories and generate explanations to mimic a content-moderation simulation. We also perform a small-scale human evaluation to measure the efficacy of these generations. An error analysis of model predictions shows an interesting correlation when compared with the predicted categories.

\newpage

\section*{Limitations}
%\jiarui{Shorten the limitations/mention harmless points}
% \paragraph{Language and Data Limitations} Our study is constrained by its exclusive use of English-language tweets, which limits the generalizability of our findings to non-English-speaking contexts. This is especially relevant for countries like Turkey, Russia, and India, where much of the local discourse occurs in other languages. As a result, our model may fail to capture the full range of censorship dynamics, including language-specific nuances and culturally embedded discourse. Furthermore, the dataset used is a 1\% subsample of Twitter, which may not fully represent the complete spectrum of censorship activities in these regions.

\paragraph{Language and Data Considerations}
Our study primarily analyzes English-language tweets, allowing for a consistent cross-country comparison. While this approach facilitates direct analysis across regions, it may not fully capture localized discourse patterns in multilingual countries such as Turkey, Russia, and India.  Furthermore, the dataset used is a 1\% subsample of Twitter, which may not fully represent the complete spectrum of censorship activities in these regions.

% \paragraph{Model Biases} LLMs, such as those employed in this study, are susceptible to biases inherited from their training data. Despite our focus on fairness and accuracy, these models may develop unintended biases toward certain cultural or political narratives. In particular, the models may overemphasize certain entities or expressions while neglecting others that carry more localized or subtle meanings. Such biases can result in skewed predictions, potentially missing less explicit but equally important examples of censored content.

\paragraph{Model Considerations}
As with all large language models, those used in this study are influenced by the characteristics of their training data. While our approach prioritizes fairness and accuracy, certain cultural or contextual variations may affect model predictions. In particular, some expressions may be more prominently recognized than others, reflecting broader patterns in the training data. These factors could shape model outputs in nuanced ways, especially when identifying less explicit forms of censored content.

% \paragraph{Temporal Constraints} Our dataset spans from 2011 to 2020, which introduces a temporal limitation as censorship policies evolve rapidly, particularly in politically sensitive regions. This means that our findings may not fully reflect the current state of censorship practices or newer moderation mechanisms that have been implemented post-2020. 

\paragraph{Temporal Scope}
Our dataset covers the period from 2011 to 2020, providing a historical perspective on censorship trends. While this enables an in-depth analysis of long-term patterns, it may not capture more recent developments in moderation practices introduced after 2020, particularly in dynamic political contexts.

% \paragraph{Platform-Specific Generalization} Our analysis focuses solely on Twitter, which has unique moderation policies and user demographics. As such, the results of our models and the observed censorship patterns may not be directly transferable to other platforms like Facebook, Instagram, or regionally specific networks. These platforms may have different guidelines, user behaviors, and censorship mechanisms, which could alter the applicability of our findings.

\paragraph{Platform-Specific Scope}
Our analysis is based on Twitter, a platform with distinct moderation policies and user demographics. While this provides valuable insights into censorship dynamics, different platforms—such as Facebook, Instagram, or regionally specific networks—may have varying guidelines and user behaviors. As a result, findings from this study primarily reflect patterns observed on Twitter, with potential variations across other platforms.

\section*{Ethical Considerations}

\paragraph{Implications of Applying LLMs to Real-World Content Moderation}
We experimented with two models from two leading companies, DeepSeek and OpenAI, on a publicly available dataset about political censorship in China across various categories, totaling 1360 queries.\footnote{\url{https://huggingface.co/datasets/promptfoo/CCP-sensitive-prompts}} We used DeepSeek R1 \citep{guo2025deepseek}, distilled on Qwen 32B \citep{bai2023qwen} and officially released on HuggingFace,\footnote{\url{https://huggingface.co/deepseek-ai/DeepSeek-R1-Distill-Qwen-32B}} alongside GPT-4o-mini \citep{gpt4o-openai} for comparison. Our results indicate that the DeepSeek model refuses 47\% of the queries, whereas the GPT model refuses 41\%. Specifically, the DeepSeek model censors more queries related to historical political movements and riots, political parties, leadership, and human rights. In contrast, GPT-4o-mini censors more queries concerning religious freedom, forced organ harvesting, and organized crime.

These findings reveal several important implications. First, the differences in refusal rates highlight that LLMs, even when trained on publicly available datasets, exhibit divergent content moderation behaviors. This suggests that moderation policies are shaped not only by the country of origin but also by company-specific guidelines, training data, and alignment strategies. Such variability raises concerns about the consistency and reliability of LLM-based moderation across different platforms. Second, the distinct censorship patterns observed underscore how LLMs may inadvertently encode biases reflecting political, cultural, or corporate priorities. These biases can shape public discourse, leading to differential access to information depending on the system in use.

\paragraph{Nature of our work} We do not propose or introduce frameworks ready for deployment, rather, we use and analyze existing tools and datasets. The classifiers may develop bias towards certain cultures or minority groups, which we have also discussed in our results. The nature of our work is "investigative journalism", due to which we have to make certain hypotheses that may or may not align with the majority views. We want to highlight that we do not hold any bias against any group mentioned in our work, but we simply make inferences based on our experiments and data. There is a potential bias from using only Twitter data, particularly given our small sample size.

% \paragraph{Political leaning.} We do not hold any political side on any of the topics discussed in this work. Due to the nature of our large-scale cross-country study, certain entities may have been targeted and we want to highlight that we remain neutral towards all such entities and do not make any targeted inferences. We simply reflect what the experiments and data showed us, and we do not hold any political views.

\paragraph{Usage of LLMs, Bias and Fairness} The generations of LLMs may not be faithful but aligned to human preferences, as has been shown by previous work \cite{turpin23-llm-unfaithful}.
%There may be certain biases that LLM generations may develop against certain groups or cultures during their training, and we, as the authors, do not share such negative sentiments.
LLMs are trained on vast amounts of data, which often contain societal biases. As a result, the models may amplify or perpetuate stereotypes and biases against certain groups based on race, gender, religion, or political affiliation. While our study focuses on detecting censorship, it is important to acknowledge that the models themselves may reflect the biases of the data they are trained on, which could disproportionately affect marginalized or minority communities. We have taken steps to mitigate these biases through careful evaluation and human oversight, but further research is needed to refine these approaches and ensure that the models are equitable in their decision-making processes.

% write a very long section

% - We do not hold political side on the topics mentioned

% - The generations of LLMs may not be faithful but aligned to human preferences, as has been shown by previous work.

% - For highly sensitive issues, we are not introducing something ready for deploy to automatically handle them

% - We are not serving surveillance, but the nature of work is more "investigative journalism"
% LLMs for investigative journalism: how are censorship decisions made?

% - The classifier might have bias towards certain culture, which we also discussed in our results, about error types.

% - There is a potential bias from using only Twitter data, particularly given our small sample size.

% - The datasets for India and Russia contain fewer tweets compared to other countries.

% - Our analysis only includes five countries, and all data is in English. We do not include tweets in German, Turkish, Hindi, Russian, or French in our analysis.

% - Come up with more :)
% E.g., \url{https://arxiv.org/pdf/2305.08283}

\paragraph{Responsible Use of Censorship Detection Models} The application of NLP models in censorship detection has significant ethical implications. While automated models can assist in identifying patterns of censorship and explaining moderation decisions, they must not be seen as a definitive solution. Censorship is a sensitive topic, particularly in politically charged or authoritarian contexts, and the outputs of these models could be misinterpreted or used to justify harmful moderation practices. We emphasize that our work should not be used as a tool for mass surveillance or to enable repressive censorship regimes. The responsibility for using these tools must lie with institutions committed to transparency, human rights, and the protection of freedom of speech.

\paragraph{Data Privacy and Consent} Our research utilizes public data from Twitter, a platform where users are often unaware of how their content may be used for academic or commercial research. While the data we analyzed is publicly available, issues surrounding the privacy and consent of users must be considered. Twitter data can sometimes include personal, sensitive, or contextually revealing information that users may not intend to be used in research, particularly studies on censorship. Researchers must be vigilant in anonymizing and safeguarding user identities where applicable, to minimize potential harm.

% \paragraph{Transparency of Automated Decisions} The use of explainability techniques such as SHAP and LIME in our work is an effort to provide transparency to the otherwise opaque decision-making process of LLMs. However, it is important to note that even these methods have limitations and may not provide a complete or fully accurate explanation of model behavior. In the context of censorship, this lack of full transparency can have serious ethical consequences if false positives or negatives in censorship prediction lead to the suppression of legitimate content or the promotion of harmful material. We advocate for the careful use of automated censorship detection systems, supplemented by human oversight, to ensure that moderation decisions are made fairly and are open to scrutiny.

\paragraph{Political and Cultural Sensitivity} Censorship is highly context-dependent, varying widely across political regimes and cultural norms. Our analysis spans five countries, each with distinct political environments and legal frameworks surrounding free speech. We acknowledge the sensitivity of this topic, particularly in regions where government control over online discourse is strong. We do not take a political stance on any of the cases or countries analyzed in this work, and we emphasize that our findings are purely academic, and aimed at advancing the understanding of how NLP models interact with censorship. The potential for misuse of our findings for political purposes or to justify further censorship remains a concern, and we strongly discourage such applications.

\ifarxiv
\section*{Acknowledgments}
This material is based in part upon work supported by the German Federal Ministry of Education and Research (BMBF): Tübingen AI Center, FKZ: 01IS18039B; by the Machine Learning Cluster of Excellence, EXC number 2064/1 – Project number 390727645. 
% MPI funding: https://atlas.is.localnet/confluence/display/SCO/Affiliation+and+Acknowledgements+in+Publications#AffiliationandAcknowledgementsinPublications-ClusterofExcellenceMachineLearning:NewPerspectivesforScience,UniversityofT%C3%BCbingen
% by a National Science Foundation award (\#2306372); by a Swiss National Science Foundation award (\#201009) and a Responsible AI grant by the Haslerstiftung.
The usage of OpenAI credits is largely supported by the Tübingen AI Center.
% Zhijing Jin is supported by PhD fellowships from the Future of Life Institute and Open Philanthropy, as well as the travel support from ELISE (GA \#951847) for the ELLIS program. 
\fi

% We are thankful to Professor Max Tegmark for his insightful discussions which inspires the early stage of the project. We would also like to thank all of our emergency annotators who were able to respond such positively in such a short time! We are also thankful to everyone that aided us for our human evaluation!
% This material is based in part upon work supported by the German Federal Ministry of Education and Research (BMBF): Tübingen AI Center, FKZ: 01IS18039B; by the Machine Learning Cluster of Excellence, EXC number 2064/1 – Project number 390727645; and
% by a National Science Foundation award (\#2306372).
%% ; by a Swiss National Science Foundation award (\#201009) and a Responsible AI grant by the Haslerstiftung.

\bibliography{custom}

\begin{thebibliography}{59}
\providecommand{\natexlab}[1]{#1}

\bibitem[{Abdelberi et~al.(2014)Abdelberi, Chen, Cunche, Cristofaro, Friedman, and K{\^{a}}afar}]{censorshipinthewild-imc14}
Chaabane Abdelberi, Terence Chen, Mathieu Cunche, Emiliano~De Cristofaro, Arik Friedman, and Mohamed~Ali K{\^{a}}afar. 2014.
\newblock \href {https://doi.org/10.1145/2663716.2663720} {Censorship in the wild: Analyzing internet filtering in syria}.
\newblock In \emph{Proceedings of the 2014 Internet Measurement Conference, {IMC} 2014, Vancouver, BC, Canada, November 5-7, 2014}, pages 285--298. {ACM}.

\bibitem[{Aiyar(2007)}]{pol-rel-india}
Mani~Shankar Aiyar. 2007.
\newblock \href {http://www.jstor.org/stable/23006045} {Politics and religion in india}.
\newblock \emph{India International Centre Quarterly}, 34(1):42--50.

\bibitem[{Akg\"{u}l and K\i{}rl\i{}do\u{g}(2015)}]{Akgul2015Internet}
Mustafa Akg\"{u}l and Melih K\i{}rl\i{}do\u{g}. 2015.
\newblock \href {https://doi.org/10.14763/2015.2.366} {Internet censorship in turkey}.
\newblock \emph{Internet Policy Review}, 4(2):1--22.

\bibitem[{Arora et~al.(2023)Arora, Nakov, Hardalov, Sarwar, Nayak, Dinkov, Zlatkova, Dent, Bhatawdekar, Bouchard et~al.}]{arora2023detecting}
Arnav Arora, Preslav Nakov, Momchil Hardalov, Sheikh~Muhammad Sarwar, Vibha Nayak, Yoan Dinkov, Dimitrina Zlatkova, Kyle Dent, Ameya Bhatawdekar, Guillaume Bouchard, et~al. 2023.
\newblock Detecting harmful content on online platforms: what platforms need vs. where research efforts go.
\newblock \emph{ACM Computing Surveys}, 56(3):1--17.

\bibitem[{Aryabumi et~al.(2024)Aryabumi, Dang, Talupuru, Dash, Cairuz, Lin, Venkitesh, Smith, Campos, Tan, Marchisio, Bartolo, Ruder, Locatelli, Kreutzer, Frosst, Gomez, Blunsom, Fadaee, Üstün, and Hooker}]{aryabumi2024aya23openweight}
Viraat Aryabumi, John Dang, Dwarak Talupuru, Saurabh Dash, David Cairuz, Hangyu Lin, Bharat Venkitesh, Madeline Smith, Jon~Ander Campos, Yi~Chern Tan, Kelly Marchisio, Max Bartolo, Sebastian Ruder, Acyr Locatelli, Julia Kreutzer, Nick Frosst, Aidan Gomez, Phil Blunsom, Marzieh Fadaee, Ahmet Üstün, and Sara Hooker. 2024.
\newblock \href {https://arxiv.org/abs/2405.15032} {Aya 23: Open weight releases to further multilingual progress}.
\newblock \emph{Preprint}, arXiv:2405.15032.

\bibitem[{Bai et~al.(2023)Bai, Bai, Chu, Cui, Dang, Deng, Fan, Ge, Han, Huang et~al.}]{bai2023qwen}
Jinze Bai, Shuai Bai, Yunfei Chu, Zeyu Cui, Kai Dang, Xiaodong Deng, Yang Fan, Wenbin Ge, Yu~Han, Fei Huang, et~al. 2023.
\newblock Qwen technical report.
\newblock \emph{arXiv preprint arXiv:2309.16609}.

\bibitem[{Belrose et~al.(2023)Belrose, Furman, Smith, Halawi, Ostrovsky, McKinney, Biderman, and Steinhardt}]{belrose2023eliciting}
Nora Belrose, Zach Furman, Logan Smith, Danny Halawi, Igor Ostrovsky, Lev McKinney, Stella Biderman, and Jacob Steinhardt. 2023.
\newblock \href {https://arxiv.org/abs/2303.08112} {Eliciting latent predictions from transformers with the tuned lens}.
\newblock \emph{Preprint}, arXiv:2303.08112.

\bibitem[{Bhargava et~al.(2021)Bhargava, Drozd, and Rogers}]{bhargava2021generalization}
Prajjwal Bhargava, Aleksandr Drozd, and Anna Rogers. 2021.
\newblock \href {https://arxiv.org/abs/2110.01518} {Generalization in nli: Ways (not) to go beyond simple heuristics}.
\newblock \emph{Preprint}, arXiv:2110.01518.

\bibitem[{Biderman et~al.(2023)Biderman, Schoelkopf, Anthony, Bradley, O’Brien, Hallahan, Khan, Purohit, Prashanth, Raff et~al.}]{biderman2023pythia}
Stella Biderman, Hailey Schoelkopf, Quentin~Gregory Anthony, Herbie Bradley, Kyle O’Brien, Eric Hallahan, Mohammad~Aflah Khan, Shivanshu Purohit, USVSN~Sai Prashanth, Edward Raff, et~al. 2023.
\newblock Pythia: A suite for analyzing large language models across training and scaling.
\newblock In \emph{International Conference on Machine Learning}, pages 2397--2430. PMLR.

\bibitem[{Bills et~al.(2023)Bills, Cammarata, Mossing, Tillman, Gao, Goh, Sutskever, Leike, Wu, and Saunders}]{bills2023language}
Steven Bills, Nick Cammarata, Dan Mossing, Henk Tillman, Leo Gao, Gabriel Goh, Ilya Sutskever, Jan Leike, Jeff Wu, and William Saunders. 2023.
\newblock Language models can explain neurons in language models.
\newblock \url{https://openaipublic.blob.core.windows.net/neuron-explainer/paper/index.html}.

\bibitem[{Boyer(2003)}]{Boyer_2003}
Dominic Boyer. 2003.
\newblock \href {https://doi.org/10.1017/S0010417503000240} {Censorship as a vocation: The institutions, practices, and cultural logic of media control in the german democratic republic}.
\newblock \emph{Comparative Studies in Society and History}, 45(3):511–545.

\bibitem[{Conmy et~al.(2023)Conmy, Mavor-Parker, Lynch, Heimersheim, and Garriga-Alonso}]{conmy23-circuit_discovery}
Arthur Conmy, Augustine Mavor-Parker, Aengus Lynch, Stefan Heimersheim, and Adri\`{a} Garriga-Alonso. 2023.
\newblock \href {https://proceedings.neurips.cc/paper_files/paper/2023/file/34e1dbe95d34d7ebaf99b9bcaeb5b2be-Paper-Conference.pdf} {Towards automated circuit discovery for mechanistic interpretability}.
\newblock In \emph{Advances in Neural Information Processing Systems}, volume~36, pages 16318--16352. Curran Associates, Inc.

\bibitem[{Conneau et~al.(2020)Conneau, Khandelwal, Goyal, Chaudhary, Wenzek, Guzm{\'a}n, Grave, Ott, Zettlemoyer, and Stoyanov}]{conneau-etal-2020-xlmr}
Alexis Conneau, Kartikay Khandelwal, Naman Goyal, Vishrav Chaudhary, Guillaume Wenzek, Francisco Guzm{\'a}n, Edouard Grave, Myle Ott, Luke Zettlemoyer, and Veselin Stoyanov. 2020.
\newblock \href {https://doi.org/10.18653/v1/2020.acl-main.747} {Unsupervised cross-lingual representation learning at scale}.
\newblock In \emph{Proceedings of the 58th Annual Meeting of the Association for Computational Linguistics}, pages 8440--8451, Online. Association for Computational Linguistics.

\bibitem[{Dalvi et~al.(2019)Dalvi, Durrani, Sajjad, Belinkov, Bau, and Glass}]{dalvi2019-whatisonegrain}
Fahim Dalvi, Nadir Durrani, Hassan Sajjad, Yonatan Belinkov, Anthony Bau, and James Glass. 2019.
\newblock \href {https://doi.org/10.1609/aaai.v33i01.33016309} {What is one grain of sand in the desert? analyzing individual neurons in deep nlp models}.
\newblock \emph{Proceedings of the AAAI Conference on Artificial Intelligence}, 33(01):6309--6317.

\bibitem[{Devlin et~al.(2018)Devlin, Chang, Lee, and Toutanova}]{devin18-bert}
Jacob Devlin, Ming{-}Wei Chang, Kenton Lee, and Kristina Toutanova. 2018.
\newblock \href {https://arxiv.org/abs/1810.04805} {{BERT:} pre-training of deep bidirectional transformers for language understanding}.
\newblock \emph{CoRR}, abs/1810.04805.

\bibitem[{Dubey et~al.(2024)Dubey, Jauhri, Pandey, Kadian, Al-Dahle, Letman, Mathur, Schelten, Yang, Fan, Goyal, Hartshorn, Yang, Mitra, Sravankumar, Korenev, Hinsvark, Rao, Zhang, Rodriguez, Gregerson, Spataru, Roziere, Biron, Tang, Chern, Caucheteux, Nayak, Bi, Marra, McConnell, Keller, Touret, Wu, Wong, Ferrer, Nikolaidis, Allonsius, Song, Pintz, Livshits, Esiobu, Choudhary, Mahajan, Garcia-Olano, Perino, Hupkes, Lakomkin, AlBadawy, Lobanova, Dinan, Smith, Radenovic, Zhang, Synnaeve, Lee, Anderson, Nail, Mialon, Pang, Cucurell, Nguyen, Korevaar, Xu, Touvron, Zarov, Ibarra, Kloumann, Misra, Evtimov, Copet, Lee, Geffert, Vranes, Park, Mahadeokar, Shah, van~der Linde, Billock, Hong, Lee, Fu, Chi, Huang, Liu, Wang, Yu, Bitton, Spisak, Park, Rocca, Johnstun, Saxe, Jia, Alwala, Upasani, Plawiak, Li, Heafield, Stone, El-Arini, Iyer, Malik, Chiu, Bhalla, Rantala-Yeary, van~der Maaten, Chen, Tan, Jenkins, Martin, Madaan, Malo, Blecher, Landzaat, de~Oliveira, Muzzi, Pasupuleti, Singh, Paluri, Kardas, Oldham, Rita,
  Pavlova, Kambadur, Lewis, Si, Singh, Hassan, Goyal, Torabi, Bashlykov, Bogoychev, Chatterji, Duchenne, Çelebi, Alrassy, Zhang, Li, Vasic, Weng, Bhargava, Dubal, Krishnan, Koura, Xu, He, Dong, Srinivasan, Ganapathy, Calderer, Cabral, Stojnic, Raileanu, Girdhar, Patel, Sauvestre, Polidoro, Sumbaly, Taylor, Silva, Hou, Wang, Hosseini, Chennabasappa, Singh, Bell, Kim, Edunov, Nie, Narang, Raparthy, Shen, Wan, Bhosale, Zhang, Vandenhende, Batra, Whitman, Sootla, Collot, Gururangan, Borodinsky, Herman, Fowler, Sheasha, Georgiou, Scialom, Speckbacher, Mihaylov, Xiao, Karn, Goswami, Gupta, Ramanathan, Kerkez, Gonguet, Do, Vogeti, Petrovic, Chu, Xiong, Fu, Meers, Martinet, Wang, Tan, Xie, Jia, Wang, Goldschlag, Gaur, Babaei, Wen, Song, Zhang, Li, Mao, Coudert, Yan, Chen, Papakipos, Singh, Grattafiori, Jain, Kelsey, Shajnfeld, Gangidi, Victoria, Goldstand, Menon, Sharma, Boesenberg, Vaughan, Baevski, Feinstein, Kallet, Sangani, Yunus, Lupu, Alvarado, Caples, Gu, Ho, Poulton, Ryan, Ramchandani, Franco, Saraf,
  Chowdhury, Gabriel, Bharambe, Eisenman, Yazdan, James, Maurer, Leonhardi, Huang, Loyd, Paola, Paranjape, Liu, Wu, Ni, Hancock, Wasti, Spence, Stojkovic, Gamido, Montalvo, Parker, Burton, Mejia, Wang, Kim, Zhou, Hu, Chu, Cai, Tindal, Feichtenhofer, Civin, Beaty, Kreymer, Li, Wyatt, Adkins, Xu, Testuggine, David, Parikh, Liskovich, Foss, Wang, Le, Holland, Dowling, Jamil, Montgomery, Presani, Hahn, Wood, Brinkman, Arcaute, Dunbar, Smothers, Sun, Kreuk, Tian, Ozgenel, Caggioni, Guzmán, Kanayet, Seide, Florez, Schwarz, Badeer, Swee, Halpern, Thattai, Herman, Sizov, Guangyi, Zhang, Lakshminarayanan, Shojanazeri, Zou, Wang, Zha, Habeeb, Rudolph, Suk, Aspegren, Goldman, Damlaj, Molybog, Tufanov, Veliche, Gat, Weissman, Geboski, Kohli, Asher, Gaya, Marcus, Tang, Chan, Zhen, Reizenstein, Teboul, Zhong, Jin, Yang, Cummings, Carvill, Shepard, McPhie, Torres, Ginsburg, Wang, Wu, U, Saxena, Prasad, Khandelwal, Zand, Matosich, Veeraraghavan, Michelena, Li, Huang, Chawla, Lakhotia, Huang, Chen, Garg, A, Silva, Bell,
  Zhang, Guo, Yu, Moshkovich, Wehrstedt, Khabsa, Avalani, Bhatt, Tsimpoukelli, Mankus, Hasson, Lennie, Reso, Groshev, Naumov, Lathi, Keneally, Seltzer, Valko, Restrepo, Patel, Vyatskov, Samvelyan, Clark, Macey, Wang, Hermoso, Metanat, Rastegari, Bansal, Santhanam, Parks, White, Bawa, Singhal, Egebo, Usunier, Laptev, Dong, Zhang, Cheng, Chernoguz, Hart, Salpekar, Kalinli, Kent, Parekh, Saab, Balaji, Rittner, Bontrager, Roux, Dollar, Zvyagina, Ratanchandani, Yuvraj, Liang, Alao, Rodriguez, Ayub, Murthy, Nayani, Mitra, Li, Hogan, Battey, Wang, Maheswari, Howes, Rinott, Bondu, Datta, Chugh, Hunt, Dhillon, Sidorov, Pan, Verma, Yamamoto, Ramaswamy, Lindsay, Lindsay, Feng, Lin, Zha, Shankar, Zhang, Zhang, Wang, Agarwal, Sajuyigbe, Chintala, Max, Chen, Kehoe, Satterfield, Govindaprasad, Gupta, Cho, Virk, Subramanian, Choudhury, Goldman, Remez, Glaser, Best, Kohler, Robinson, Li, Zhang, Matthews, Chou, Shaked, Vontimitta, Ajayi, Montanez, Mohan, Kumar, Mangla, Albiero, Ionescu, Poenaru, Mihailescu, Ivanov, Li, Wang,
  Jiang, Bouaziz, Constable, Tang, Wang, Wu, Wang, Xia, Wu, Gao, Chen, Hu, Jia, Qi, Li, Zhang, Zhang, Adi, Nam, Yu, Wang, Hao, Qian, He, Rait, DeVito, Rosnbrick, Wen, Yang, and Zhao}]{dubey2024llama3herdmodels}
Abhimanyu Dubey, Abhinav Jauhri, Abhinav Pandey, Abhishek Kadian, Ahmad Al-Dahle, Aiesha Letman, Akhil Mathur, Alan Schelten, Amy Yang, Angela Fan, Anirudh Goyal, Anthony Hartshorn, Aobo Yang, Archi Mitra, Archie Sravankumar, Artem Korenev, Arthur Hinsvark, Arun Rao, Aston Zhang, Aurelien Rodriguez, Austen Gregerson, Ava Spataru, Baptiste Roziere, Bethany Biron, Binh Tang, Bobbie Chern, Charlotte Caucheteux, Chaya Nayak, Chloe Bi, Chris Marra, Chris McConnell, Christian Keller, Christophe Touret, Chunyang Wu, Corinne Wong, Cristian~Canton Ferrer, Cyrus Nikolaidis, Damien Allonsius, Daniel Song, Danielle Pintz, Danny Livshits, David Esiobu, Dhruv Choudhary, Dhruv Mahajan, Diego Garcia-Olano, Diego Perino, Dieuwke Hupkes, Egor Lakomkin, Ehab AlBadawy, Elina Lobanova, Emily Dinan, Eric~Michael Smith, Filip Radenovic, Frank Zhang, Gabriel Synnaeve, Gabrielle Lee, Georgia~Lewis Anderson, Graeme Nail, Gregoire Mialon, Guan Pang, Guillem Cucurell, Hailey Nguyen, Hannah Korevaar, Hu~Xu, Hugo Touvron, Iliyan Zarov,
  Imanol~Arrieta Ibarra, Isabel Kloumann, Ishan Misra, Ivan Evtimov, Jade Copet, Jaewon Lee, Jan Geffert, Jana Vranes, Jason Park, Jay Mahadeokar, Jeet Shah, Jelmer van~der Linde, Jennifer Billock, Jenny Hong, Jenya Lee, Jeremy Fu, Jianfeng Chi, Jianyu Huang, Jiawen Liu, Jie Wang, Jiecao Yu, Joanna Bitton, Joe Spisak, Jongsoo Park, Joseph Rocca, Joshua Johnstun, Joshua Saxe, Junteng Jia, Kalyan~Vasuden Alwala, Kartikeya Upasani, Kate Plawiak, Ke~Li, Kenneth Heafield, Kevin Stone, Khalid El-Arini, Krithika Iyer, Kshitiz Malik, Kuenley Chiu, Kunal Bhalla, Lauren Rantala-Yeary, Laurens van~der Maaten, Lawrence Chen, Liang Tan, Liz Jenkins, Louis Martin, Lovish Madaan, Lubo Malo, Lukas Blecher, Lukas Landzaat, Luke de~Oliveira, Madeline Muzzi, Mahesh Pasupuleti, Mannat Singh, Manohar Paluri, Marcin Kardas, Mathew Oldham, Mathieu Rita, Maya Pavlova, Melanie Kambadur, Mike Lewis, Min Si, Mitesh~Kumar Singh, Mona Hassan, Naman Goyal, Narjes Torabi, Nikolay Bashlykov, Nikolay Bogoychev, Niladri Chatterji, Olivier
  Duchenne, Onur Çelebi, Patrick Alrassy, Pengchuan Zhang, Pengwei Li, Petar Vasic, Peter Weng, Prajjwal Bhargava, Pratik Dubal, Praveen Krishnan, Punit~Singh Koura, Puxin Xu, Qing He, Qingxiao Dong, Ragavan Srinivasan, Raj Ganapathy, Ramon Calderer, Ricardo~Silveira Cabral, Robert Stojnic, Roberta Raileanu, Rohit Girdhar, Rohit Patel, Romain Sauvestre, Ronnie Polidoro, Roshan Sumbaly, Ross Taylor, Ruan Silva, Rui Hou, Rui Wang, Saghar Hosseini, Sahana Chennabasappa, Sanjay Singh, Sean Bell, Seohyun~Sonia Kim, Sergey Edunov, Shaoliang Nie, Sharan Narang, Sharath Raparthy, Sheng Shen, Shengye Wan, Shruti Bhosale, Shun Zhang, Simon Vandenhende, Soumya Batra, Spencer Whitman, Sten Sootla, Stephane Collot, Suchin Gururangan, Sydney Borodinsky, Tamar Herman, Tara Fowler, Tarek Sheasha, Thomas Georgiou, Thomas Scialom, Tobias Speckbacher, Todor Mihaylov, Tong Xiao, Ujjwal Karn, Vedanuj Goswami, Vibhor Gupta, Vignesh Ramanathan, Viktor Kerkez, Vincent Gonguet, Virginie Do, Vish Vogeti, Vladan Petrovic, Weiwei Chu,
  Wenhan Xiong, Wenyin Fu, Whitney Meers, Xavier Martinet, Xiaodong Wang, Xiaoqing~Ellen Tan, Xinfeng Xie, Xuchao Jia, Xuewei Wang, Yaelle Goldschlag, Yashesh Gaur, Yasmine Babaei, Yi~Wen, Yiwen Song, Yuchen Zhang, Yue Li, Yuning Mao, Zacharie~Delpierre Coudert, Zheng Yan, Zhengxing Chen, Zoe Papakipos, Aaditya Singh, Aaron Grattafiori, Abha Jain, Adam Kelsey, Adam Shajnfeld, Adithya Gangidi, Adolfo Victoria, Ahuva Goldstand, Ajay Menon, Ajay Sharma, Alex Boesenberg, Alex Vaughan, Alexei Baevski, Allie Feinstein, Amanda Kallet, Amit Sangani, Anam Yunus, Andrei Lupu, Andres Alvarado, Andrew Caples, Andrew Gu, Andrew Ho, Andrew Poulton, Andrew Ryan, Ankit Ramchandani, Annie Franco, Aparajita Saraf, Arkabandhu Chowdhury, Ashley Gabriel, Ashwin Bharambe, Assaf Eisenman, Azadeh Yazdan, Beau James, Ben Maurer, Benjamin Leonhardi, Bernie Huang, Beth Loyd, Beto~De Paola, Bhargavi Paranjape, Bing Liu, Bo~Wu, Boyu Ni, Braden Hancock, Bram Wasti, Brandon Spence, Brani Stojkovic, Brian Gamido, Britt Montalvo, Carl
  Parker, Carly Burton, Catalina Mejia, Changhan Wang, Changkyu Kim, Chao Zhou, Chester Hu, Ching-Hsiang Chu, Chris Cai, Chris Tindal, Christoph Feichtenhofer, Damon Civin, Dana Beaty, Daniel Kreymer, Daniel Li, Danny Wyatt, David Adkins, David Xu, Davide Testuggine, Delia David, Devi Parikh, Diana Liskovich, Didem Foss, Dingkang Wang, Duc Le, Dustin Holland, Edward Dowling, Eissa Jamil, Elaine Montgomery, Eleonora Presani, Emily Hahn, Emily Wood, Erik Brinkman, Esteban Arcaute, Evan Dunbar, Evan Smothers, Fei Sun, Felix Kreuk, Feng Tian, Firat Ozgenel, Francesco Caggioni, Francisco Guzmán, Frank Kanayet, Frank Seide, Gabriela~Medina Florez, Gabriella Schwarz, Gada Badeer, Georgia Swee, Gil Halpern, Govind Thattai, Grant Herman, Grigory Sizov, Guangyi, Zhang, Guna Lakshminarayanan, Hamid Shojanazeri, Han Zou, Hannah Wang, Hanwen Zha, Haroun Habeeb, Harrison Rudolph, Helen Suk, Henry Aspegren, Hunter Goldman, Ibrahim Damlaj, Igor Molybog, Igor Tufanov, Irina-Elena Veliche, Itai Gat, Jake Weissman, James
  Geboski, James Kohli, Japhet Asher, Jean-Baptiste Gaya, Jeff Marcus, Jeff Tang, Jennifer Chan, Jenny Zhen, Jeremy Reizenstein, Jeremy Teboul, Jessica Zhong, Jian Jin, Jingyi Yang, Joe Cummings, Jon Carvill, Jon Shepard, Jonathan McPhie, Jonathan Torres, Josh Ginsburg, Junjie Wang, Kai Wu, Kam~Hou U, Karan Saxena, Karthik Prasad, Kartikay Khandelwal, Katayoun Zand, Kathy Matosich, Kaushik Veeraraghavan, Kelly Michelena, Keqian Li, Kun Huang, Kunal Chawla, Kushal Lakhotia, Kyle Huang, Lailin Chen, Lakshya Garg, Lavender A, Leandro Silva, Lee Bell, Lei Zhang, Liangpeng Guo, Licheng Yu, Liron Moshkovich, Luca Wehrstedt, Madian Khabsa, Manav Avalani, Manish Bhatt, Maria Tsimpoukelli, Martynas Mankus, Matan Hasson, Matthew Lennie, Matthias Reso, Maxim Groshev, Maxim Naumov, Maya Lathi, Meghan Keneally, Michael~L. Seltzer, Michal Valko, Michelle Restrepo, Mihir Patel, Mik Vyatskov, Mikayel Samvelyan, Mike Clark, Mike Macey, Mike Wang, Miquel~Jubert Hermoso, Mo~Metanat, Mohammad Rastegari, Munish Bansal, Nandhini
  Santhanam, Natascha Parks, Natasha White, Navyata Bawa, Nayan Singhal, Nick Egebo, Nicolas Usunier, Nikolay~Pavlovich Laptev, Ning Dong, Ning Zhang, Norman Cheng, Oleg Chernoguz, Olivia Hart, Omkar Salpekar, Ozlem Kalinli, Parkin Kent, Parth Parekh, Paul Saab, Pavan Balaji, Pedro Rittner, Philip Bontrager, Pierre Roux, Piotr Dollar, Polina Zvyagina, Prashant Ratanchandani, Pritish Yuvraj, Qian Liang, Rachad Alao, Rachel Rodriguez, Rafi Ayub, Raghotham Murthy, Raghu Nayani, Rahul Mitra, Raymond Li, Rebekkah Hogan, Robin Battey, Rocky Wang, Rohan Maheswari, Russ Howes, Ruty Rinott, Sai~Jayesh Bondu, Samyak Datta, Sara Chugh, Sara Hunt, Sargun Dhillon, Sasha Sidorov, Satadru Pan, Saurabh Verma, Seiji Yamamoto, Sharadh Ramaswamy, Shaun Lindsay, Shaun Lindsay, Sheng Feng, Shenghao Lin, Shengxin~Cindy Zha, Shiva Shankar, Shuqiang Zhang, Shuqiang Zhang, Sinong Wang, Sneha Agarwal, Soji Sajuyigbe, Soumith Chintala, Stephanie Max, Stephen Chen, Steve Kehoe, Steve Satterfield, Sudarshan Govindaprasad, Sumit Gupta,
  Sungmin Cho, Sunny Virk, Suraj Subramanian, Sy~Choudhury, Sydney Goldman, Tal Remez, Tamar Glaser, Tamara Best, Thilo Kohler, Thomas Robinson, Tianhe Li, Tianjun Zhang, Tim Matthews, Timothy Chou, Tzook Shaked, Varun Vontimitta, Victoria Ajayi, Victoria Montanez, Vijai Mohan, Vinay~Satish Kumar, Vishal Mangla, Vítor Albiero, Vlad Ionescu, Vlad Poenaru, Vlad~Tiberiu Mihailescu, Vladimir Ivanov, Wei Li, Wenchen Wang, Wenwen Jiang, Wes Bouaziz, Will Constable, Xiaocheng Tang, Xiaofang Wang, Xiaojian Wu, Xiaolan Wang, Xide Xia, Xilun Wu, Xinbo Gao, Yanjun Chen, Ye~Hu, Ye~Jia, Ye~Qi, Yenda Li, Yilin Zhang, Ying Zhang, Yossi Adi, Youngjin Nam, Yu, Wang, Yuchen Hao, Yundi Qian, Yuzi He, Zach Rait, Zachary DeVito, Zef Rosnbrick, Zhaoduo Wen, Zhenyu Yang, and Zhiwei Zhao. 2024.
\newblock \href {https://arxiv.org/abs/2407.21783} {The llama 3 herd of models}.
\newblock \emph{Preprint}, arXiv:2407.21783.

\bibitem[{Elmas et~al.(2021)Elmas, Overdorf, and Aberer}]{censorship-dataset-aaai21}
Tuğrulcan Elmas, Rebekah Overdorf, and Karl Aberer. 2021.
\newblock \href {https://doi.org/10.1609/icwsm.v15i1.18124} {A dataset of state-censored tweets}.
\newblock \emph{Proceedings of the International AAAI Conference on Web and Social Media}, 15(1):1009--1015.

\bibitem[{Founta et~al.(2019)Founta, Chatzakou, Kourtellis, Blackburn, Vakali, and Leontiadis}]{founta-hatespeech}
Antigoni~Maria Founta, Despoina Chatzakou, Nicolas Kourtellis, Jeremy Blackburn, Athena Vakali, and Ilias Leontiadis. 2019.
\newblock \href {https://doi.org/10.1145/3292522.3326028} {A unified deep learning architecture for abuse detection}.
\newblock In \emph{Proceedings of the 10th ACM Conference on Web Science}, WebSci '19, page 105–114, New York, NY, USA. Association for Computing Machinery.

\bibitem[{Ghandeharioun et~al.(2024)Ghandeharioun, Caciularu, Pearce, Dixon, and Geva}]{ghandeharioun2024patchscopes}
Asma Ghandeharioun, Avi Caciularu, Adam Pearce, Lucas Dixon, and Mor Geva. 2024.
\newblock \href {https://openreview.net/forum?id=5uwBzcn885} {Patchscopes: A unifying framework for inspecting hidden representations of language models}.
\newblock In \emph{Forty-first International Conference on Machine Learning}.

\bibitem[{Ghorbani and Zou(2019)}]{ghorbani2019data}
Amirata Ghorbani and James Zou. 2019.
\newblock Data shapley: Equitable valuation of data for machine learning.
\newblock In \emph{International Conference on Machine Learning}, pages 2242--2251.

\bibitem[{Gligoric et~al.(2024)Gligoric, Cheng, Zheng, Durmus, and Jurafsky}]{gligoric-etal-2024-nlp}
Kristina Gligoric, Myra Cheng, Lucia Zheng, Esin Durmus, and Dan Jurafsky. 2024.
\newblock \href {https://doi.org/10.18653/v1/2024.naacl-long.331} {{NLP} systems that can{'}t tell use from mention censor counterspeech, but teaching the distinction helps}.
\newblock In \emph{Proceedings of the 2024 Conference of the North American Chapter of the Association for Computational Linguistics: Human Language Technologies (Volume 1: Long Papers)}, pages 5942--5959, Mexico City, Mexico. Association for Computational Linguistics.

\bibitem[{Gorwa et~al.(2020)Gorwa, Binns, and Katzenbach}]{gorwa2020algorithmic}
Robert Gorwa, Reuben Binns, and Christian Katzenbach. 2020.
\newblock Algorithmic content moderation: Technical and political challenges in the automation of platform governance.
\newblock \emph{Big Data \& Society}, 7(1):2053951719897945.

\bibitem[{Grimmelmann(2015)}]{grimmelmann2015virtues}
James Grimmelmann. 2015.
\newblock The virtues of moderation.
\newblock \emph{Yale JL \& Tech.}, 17:42.

\bibitem[{Guo et~al.(2025)Guo, Yang, Zhang, Song, Zhang, Xu, Zhu, Ma, Wang, Bi et~al.}]{guo2025deepseek}
Daya Guo, Dejian Yang, Haowei Zhang, Junxiao Song, Ruoyu Zhang, Runxin Xu, Qihao Zhu, Shirong Ma, Peiyi Wang, Xiao Bi, et~al. 2025.
\newblock Deepseek-r1: Incentivizing reasoning capability in llms via reinforcement learning.
\newblock \emph{arXiv preprint arXiv:2501.12948}.

\bibitem[{Huang(2024{\natexlab{a}})}]{huang2024content}
Tao Huang. 2024{\natexlab{a}}.
\newblock Content moderation by llm: From accuracy to legitimacy.
\newblock \emph{arXiv preprint arXiv:2409.03219}.

\bibitem[{Huang(2024{\natexlab{b}})}]{huang2024contentmoderationllmaccuracy}
Tao Huang. 2024{\natexlab{b}}.
\newblock \href {https://arxiv.org/abs/2409.03219} {Content moderation by llm: From accuracy to legitimacy}.
\newblock \emph{Preprint}, arXiv:2409.03219.

\bibitem[{Jaki and Smedt(2019)}]{jaki2019rightwing-germany}
Sylvia Jaki and Tom~De Smedt. 2019.
\newblock \href {https://arxiv.org/abs/1910.07518} {Right-wing german hate speech on twitter: Analysis and automatic detection}.
\newblock \emph{CoRR}, abs/1910.07518.

\bibitem[{Ji and Knight(2018)}]{ji-knight-2018-creative}
Heng Ji and Kevin Knight. 2018.
\newblock \href {https://aclanthology.org/W18-4203} {Creative language encoding under censorship}.
\newblock In \emph{Proceedings of the First Workshop on Natural Language Processing for {I}nternet Freedom}, pages 23--33, Santa Fe, New Mexico, USA. Association for Computational Linguistics.

\bibitem[{Knockel et~al.(2018)Knockel, Crete-Nishihata, and Ruan}]{knockel-etal-2018-effect}
Jeffrey Knockel, Masashi Crete-Nishihata, and Lotus Ruan. 2018.
\newblock \href {https://aclanthology.org/W18-4201} {The effect of information controls on developers in {C}hina: An analysis of censorship in {C}hinese open source projects}.
\newblock In \emph{Proceedings of the First Workshop on Natural Language Processing for {I}nternet Freedom}, pages 1--11, Santa Fe, New Mexico, USA. Association for Computational Linguistics.

\bibitem[{Koh and Liang(2017)}]{pmlr-v70-koh17a}
Pang~Wei Koh and Percy Liang. 2017.
\newblock \href {https://proceedings.mlr.press/v70/koh17a.html} {Understanding black-box predictions via influence functions}.
\newblock In \emph{Proceedings of the 34th International Conference on Machine Learning}, volume~70 of \emph{Proceedings of Machine Learning Research}, pages 1885--1894. PMLR.

\bibitem[{Kolla et~al.(2024)Kolla, Salunkhe, Chandrasekharan, and Saha}]{Kolla2024LLMModCL}
Mahi Kolla, Siddharth Salunkhe, Eshwar Chandrasekharan, and Koustuv Saha. 2024.
\newblock \href {https://api.semanticscholar.org/CorpusID:269101741} {Llm-mod: Can large language models assist content moderation?}
\newblock \emph{Extended Abstracts of the CHI Conference on Human Factors in Computing Systems}.

\bibitem[{Ksenia~Ermoshina and Musiani(2022)}]{russia-cens}
Benjamin~Loveluck Ksenia~Ermoshina and Francesca Musiani. 2022.
\newblock \href {https://doi.org/10.1080/19331681.2021.1905972} {A market of black boxes: The political economy of internet surveillance and censorship in russia}.
\newblock \emph{Journal of Information Technology \& Politics}, 19(1):18--33.

\bibitem[{Kumar et~al.(2024)Kumar, AbuHashem, and Durumeric}]{watchyourlanguage-icwsm24}
Deepak Kumar, Yousef~Anees AbuHashem, and Zakir Durumeric. 2024.
\newblock \href {https://doi.org/10.1609/icwsm.v18i1.31358} {Watch your language: Investigating content moderation with large language models}.
\newblock \emph{Proceedings of the International AAAI Conference on Web and Social Media}, 18(1):865--878.

\bibitem[{Kunz and Kuhlmann(2024)}]{kunz-kuhlmann-2024-properties}
Jenny Kunz and Marco Kuhlmann. 2024.
\newblock \href {https://doi.org/10.18653/v1/2024.hcinlp-1.2} {Properties and challenges of {LLM}-generated explanations}.
\newblock In \emph{Proceedings of the Third Workshop on Bridging Human--Computer Interaction and Natural Language Processing}, pages 13--27, Mexico City, Mexico. Association for Computational Linguistics.

\bibitem[{Lee et~al.(2024)Lee, Goldwasser, and Reese}]{lee-etal-2024-towards}
Younghun Lee, Dan Goldwasser, and Laura~Schwab Reese. 2024.
\newblock \href {https://aclanthology.org/2024.findings-eacl.137} {Towards understanding counseling conversations: Domain knowledge and large language models}.
\newblock In \emph{Findings of the Association for Computational Linguistics: EACL 2024}, pages 2032--2047, St. Julian{'}s, Malta. Association for Computational Linguistics.

\bibitem[{Liu et~al.(2019)Liu, Ott, Goyal, Du, Joshi, Chen, Levy, Lewis, Zettlemoyer, and Stoyanov}]{roberta-2019}
Yinhan Liu, Myle Ott, Naman Goyal, Jingfei Du, Mandar Joshi, Danqi Chen, Omer Levy, Mike Lewis, Luke Zettlemoyer, and Veselin Stoyanov. 2019.
\newblock \href {https://arxiv.org/abs/1907.11692} {Roberta: {A} robustly optimized {BERT} pretraining approach}.
\newblock \emph{CoRR}, abs/1907.11692.

\bibitem[{Lundberg and Lee(2017)}]{lundberg-shap2017}
Scott~M Lundberg and Su-In Lee. 2017.
\newblock \href {https://proceedings.neurips.cc/paper_files/paper/2017/file/8a20a8621978632d76c43dfd28b67767-Paper.pdf} {A unified approach to interpreting model predictions}.
\newblock In \emph{Advances in Neural Information Processing Systems}, volume~30. Curran Associates, Inc.

\bibitem[{MacKinnon(2009)}]{MacKinnon_2009}
Rebecca MacKinnon. 2009.
\newblock \href {https://doi.org/10.5210/fm.v14i2.2378} {China’s censorship 2.0: How companies censor bloggers}.
\newblock \emph{First Monday}, 14(2).

\bibitem[{Masud et~al.(2024)Masud, Singh, Hangya, Fraser, and Chakraborty}]{masud2024hatepersonifiedinvestigatingrole}
Sarah Masud, Sahajpreet Singh, Viktor Hangya, Alexander Fraser, and Tanmoy Chakraborty. 2024.
\newblock \href {https://arxiv.org/abs/2410.02657} {Hate personified: Investigating the role of llms in content moderation}.
\newblock \emph{Preprint}, arXiv:2410.02657.

\bibitem[{Mathew et~al.(2021)Mathew, Saha, Yimam, Biemann, Goyal, and Mukherjee}]{mathew2021hatexplain}
Binny Mathew, Punyajoy Saha, Seid~Muhie Yimam, Chris Biemann, Pawan Goyal, and Animesh Mukherjee. 2021.
\newblock Hatexplain: A benchmark dataset for explainable hate speech detection.
\newblock In \emph{Proceedings of the AAAI Conference on Artificial Intelligence}, volume~35, pages 14867--14875.

\bibitem[{Mchangama and Fiss(2019)}]{mchangama2019digital}
Jacob Mchangama and Joelle Fiss. 2019.
\newblock The digital berlin wall: How germany (accidentally) created a prototype for global online censorship.
\newblock \emph{Copenhagen: Justitia and Authors}.

\bibitem[{Meng et~al.(2022)Meng, Bau, Andonian, and Belinkov}]{meng2022-rome}
Kevin Meng, David Bau, Alex Andonian, and Yonatan Belinkov. 2022.
\newblock \href {https://proceedings.neurips.cc/paper_files/paper/2022/file/6f1d43d5a82a37e89b0665b33bf3a182-Paper-Conference.pdf} {Locating and editing factual associations in gpt}.
\newblock In \emph{Advances in Neural Information Processing Systems}, volume~35, pages 17359--17372. Curran Associates, Inc.

\bibitem[{nostalgebraist(2020)}]{nostalgebraist2020}
nostalgebraist. 2020.
\newblock \href {https://www.lesswrong.com/posts/AcKRB8wDpdaN6v6ru/interpreting-gpt-the-logit-lens} {interpreting gpt: the logit lens}.

\bibitem[{OpenAI(2024)}]{gpt4o-openai}
OpenAI. 2024.
\newblock \href {https://openai.com/index/hello-gpt-4o/} {Gpt-4o}.

\bibitem[{Paasch-Colberg et~al.(2021)Paasch-Colberg, Strippel, Trebbe, and Emmer}]{paasch2021insult}
S{\"u}nje Paasch-Colberg, Christian Strippel, Joachim Trebbe, and Martin Emmer. 2021.
\newblock From insult to hate speech: Mapping offensive language in german user comments on immigration.
\newblock \emph{Media and Communication}, 9(1):171--180.

\bibitem[{Pate(2023)}]{pate2023investigating}
Holly Pate. 2023.
\newblock \href {https://gijn.org/stories/investigating-social-media-companies/} {Experts discuss investigating social media companies like facebook, twitter, and tiktok}.
\newblock \emph{Global Investigative Journalism Network}.

\bibitem[{Rogers et~al.(2019)Rogers, Kovaleva, and Rumshisky}]{rogers-etal-2019-calls}
Anna Rogers, Olga Kovaleva, and Anna Rumshisky. 2019.
\newblock \href {https://doi.org/10.18653/v1/D19-5005} {Calls to action on social media: Detection, social impact, and censorship potential}.
\newblock In \emph{Proceedings of the Second Workshop on Natural Language Processing for Internet Freedom: Censorship, Disinformation, and Propaganda}, pages 36--44, Hong Kong, China. Association for Computational Linguistics.

\bibitem[{Schoch et~al.(2023)Schoch, Mishra, and Ji}]{schoch-etal-2023-data}
Stephanie Schoch, Ritwick Mishra, and Yangfeng Ji. 2023.
\newblock \href {https://doi.org/10.18653/v1/2023.acl-srw.37} {Data selection for fine-tuning large language models using transferred shapley values}.
\newblock In \emph{Proceedings of the 61st Annual Meeting of the Association for Computational Linguistics (Volume 4: Student Research Workshop)}, pages 266--275, Toronto, Canada. Association for Computational Linguistics.

\bibitem[{Suzor et~al.(2019)Suzor, West, Quodling, and York}]{suzor2019we}
Nicolas~P Suzor, Sarah~Myers West, Andrew Quodling, and Jillian York. 2019.
\newblock What do we mean when we talk about transparency? toward meaningful transparency in commercial content moderation.
\newblock \emph{International Journal of Communication}, 13:18.

\bibitem[{Swayamdipta et~al.(2020)Swayamdipta, Schwartz, Lourie, Wang, Hajishirzi, Smith, and Choi}]{swayamdipta-etal-2020-dataset}
Swabha Swayamdipta, Roy Schwartz, Nicholas Lourie, Yizhong Wang, Hannaneh Hajishirzi, Noah~A. Smith, and Yejin Choi. 2020.
\newblock \href {https://doi.org/10.18653/v1/2020.emnlp-main.746} {Dataset cartography: Mapping and diagnosing datasets with training dynamics}.
\newblock In \emph{Proceedings of the 2020 Conference on Empirical Methods in Natural Language Processing (EMNLP)}, pages 9275--9293, Online. Association for Computational Linguistics.

\bibitem[{Thomas et~al.(2024)Thomas, Deka, Raja, and Sathyan}]{inter-rel-pol}
Vineeth Thomas, Chandana Deka, Aparajitha Raja, and Arsha~V Sathyan. 2024.
\newblock \href {https://doi.org/10.3390/rel15121509} {Examining the intervention of religion in indian politics through hindutva under the modi regime}.
\newblock \emph{Religions}, 15(12).

\bibitem[{Thorne et~al.(2018)Thorne, Vlachos, Christodoulopoulos, and Mittal}]{thorne-etal-2018-fever}
James Thorne, Andreas Vlachos, Christos Christodoulopoulos, and Arpit Mittal. 2018.
\newblock \href {https://doi.org/10.18653/v1/N18-1074} {{FEVER}: a large-scale dataset for fact extraction and {VER}ification}.
\newblock In \emph{Proceedings of the 2018 Conference of the North {A}merican Chapter of the Association for Computational Linguistics: Human Language Technologies, Volume 1 (Long Papers)}, pages 809--819, New Orleans, Louisiana. Association for Computational Linguistics.

\bibitem[{Turpin et~al.(2023)Turpin, Michael, Perez, and Bowman}]{turpin23-llm-unfaithful}
Miles Turpin, Julian Michael, Ethan Perez, and Samuel Bowman. 2023.
\newblock \href {https://proceedings.neurips.cc/paper_files/paper/2023/file/ed3fea9033a80fea1376299fa7863f4a-Paper-Conference.pdf} {Language models don\textquotesingle t always say what they think: Unfaithful explanations in chain-of-thought prompting}.
\newblock In \emph{Advances in Neural Information Processing Systems}, volume~36, pages 74952--74965. Curran Associates, Inc.

\bibitem[{Vig et~al.(2020)Vig, Gehrmann, Belinkov, Qian, Nevo, Singer, and Shieber}]{vigetal2020-cma}
Jesse Vig, Sebastian Gehrmann, Yonatan Belinkov, Sharon Qian, Daniel Nevo, Yaron Singer, and Stuart Shieber. 2020.
\newblock \href {https://proceedings.neurips.cc/paper_files/paper/2020/file/92650b2e92217715fe312e6fa7b90d82-Paper.pdf} {Investigating gender bias in language models using causal mediation analysis}.
\newblock In \emph{Advances in Neural Information Processing Systems}, volume~33, pages 12388--12401. Curran Associates, Inc.

\bibitem[{Warf(2010)}]{Warf2010}
Barney Warf. 2010.
\newblock \href {https://doi.org/10.1007/s10708-010-9393-3} {Geographies of global internet censorship}.
\newblock \emph{GeoJournal}, 76(1):1–23.

\bibitem[{Waseem et~al.(2017)Waseem, Davidson, Warmsley, and Weber}]{waseem-etal-2017-understanding}
Zeerak Waseem, Thomas Davidson, Dana Warmsley, and Ingmar Weber. 2017.
\newblock \href {https://doi.org/10.18653/v1/W17-3012} {Understanding abuse: A typology of abusive language detection subtasks}.
\newblock In \emph{Proceedings of the First Workshop on Abusive Language Online}, pages 78--84, Vancouver, BC, Canada. Association for Computational Linguistics.

\bibitem[{Waseem and Hovy(2016)}]{waseem-hovy-2016-hateful}
Zeerak Waseem and Dirk Hovy. 2016.
\newblock \href {https://doi.org/10.18653/v1/N16-2013} {Hateful symbols or hateful people? predictive features for hate speech detection on {T}witter}.
\newblock In \emph{Proceedings of the {NAACL} Student Research Workshop}, pages 88--93, San Diego, California. Association for Computational Linguistics.

\bibitem[{Wolf et~al.(2020)Wolf, Debut, Sanh, Chaumond, Delangue, Moi, Cistac, Rault, Louf, Funtowicz, Davison, Shleifer, von Platen, Ma, Jernite, Plu, Xu, Le~Scao, Gugger, Drame, Lhoest, and Rush}]{wolf-etal-2020-transformers}
Thomas Wolf, Lysandre Debut, Victor Sanh, Julien Chaumond, Clement Delangue, Anthony Moi, Pierric Cistac, Tim Rault, Remi Louf, Morgan Funtowicz, Joe Davison, Sam Shleifer, Patrick von Platen, Clara Ma, Yacine Jernite, Julien Plu, Canwen Xu, Teven Le~Scao, Sylvain Gugger, Mariama Drame, Quentin Lhoest, and Alexander Rush. 2020.
\newblock \href {https://doi.org/10.18653/v1/2020.emnlp-demos.6} {Transformers: State-of-the-art natural language processing}.
\newblock In \emph{Proceedings of the 2020 Conference on Empirical Methods in Natural Language Processing: System Demonstrations}, pages 38--45, Online. Association for Computational Linguistics.

\bibitem[{Yadav et~al.(2024)Yadav, Masud, Goyal, Akhtar, and Chakraborty}]{yadav-etal-2024-tox}
Neemesh Yadav, Sarah Masud, Vikram Goyal, Md~Shad Akhtar, and Tanmoy Chakraborty. 2024.
\newblock \href {https://aclanthology.org/2024.findings-acl.831} {Tox-{BART}: Leveraging toxicity attributes for explanation generation of implicit hate speech}.
\newblock In \emph{Findings of the Association for Computational Linguistics ACL 2024}, pages 13967--13983, Bangkok, Thailand and virtual meeting. Association for Computational Linguistics.

\end{thebibliography}

\appendix

\section{Experimental Details}
\label{appendix:exp_details}

We performed our experiments via the Huggingface Transformers library \cite{wolf-etal-2020-transformers}. We used open-source LLMs with a commercial license such as Llama-3.1 and Llama-3.2, and completely open-source LLMs such as Aya-23 and the Pythia. All models were trained for one epoch, with a learning rate of 1e-5, a drop-out rate of 0.1, and a batch size of 8.

Our experiments were performed on NVIDIA H100 and A100 clusters, and we used the OpenAI API to query \texttt{GPT-4o-mini}.

\paragraph{LLMs used.} We used the LLMs mentioned in Table~\ref{tab:model-list}, for our zero-shot censorship classification and explainability experiments. The Llama-3.1 series is among the best-performing medium-scale open-source LLMs, while the Aya-23 suite offers one of the strongest multilingual LLMs, covering 23 languages. GPT-4o-mini performs only slightly below GPT-4o but is significantly more cost-effective, making it an ideal choice for our experiments. 

\begin{table}[ht]
\resizebox{\linewidth}{!}{
\begin{tabular}{ll}
\toprule
\textbf{Model} & \textbf{Knowledge Cutoff} \\ \toprule
Aya23-8B$^{\dagger}$ \cite{aryabumi2024aya23openweight} & Unknown \\
Llama-3.1-8B-Instruct$^{\dagger}$ \cite{dubey2024llama3herdmodels} & December 2023 \\
GPT-4o-mini$^{\ddagger}$ \cite{gpt4o-openai} & October 2023 \\
GPT-4o$^{\ddagger}$ \cite{gpt4o-openai} & October 2023 \\
 \bottomrule
\end{tabular}
}
\caption{\textbf{Details of LLMs Used in the Explanation Studies.} Dagger ($^{\dagger}$) refers to open-weight models, and Double Dagger ($^{\ddagger}$) refers to closed-source models}
\label{tab:model-list}
\vspace{-1em}
\end{table}

\begin{table}[t]
\centering
\resizebox{\linewidth}{!}{ % Ensures it fits within the column width
\normalsize
\begin{tabular}{p{4em} p{13em} p{4em} } % Adjust column widths for better text flow
\toprule
\textbf{Country} & \textbf{Censored Post} \\ \toprule
\textbf{Germany}  &  what is the sense of this lol \\ \midrule
\textbf{France} & Gaia is knocking \\ \midrule
\textbf{India}  & Another one bites the dust Another one bites the dust And another one gone and another one gone Another one bites the dust \#INDvNZL \#NZvsIND \\ \midrule
\textbf{Turkey}  & Stop focusing on how stressed you are and remember how BLESSED you are \\ \midrule
\textbf{Russia} & In Brazil even the dogs are good at football \\ \bottomrule
\end{tabular}
}
\caption{Examples of censored tweets belonging to the \texttt{Other} category according to \texttt{GPT-4o-mini}, across all countries.}

\label{tab:other-cat-ex}
\vspace{-1em}
\end{table}

\section{Censorship Categorization}
\label{appendix:category_examples}

\begin{figure*}
\centering
    \begin{subfigure}{0.48\linewidth}
        \centering
        \includegraphics[width=\linewidth]{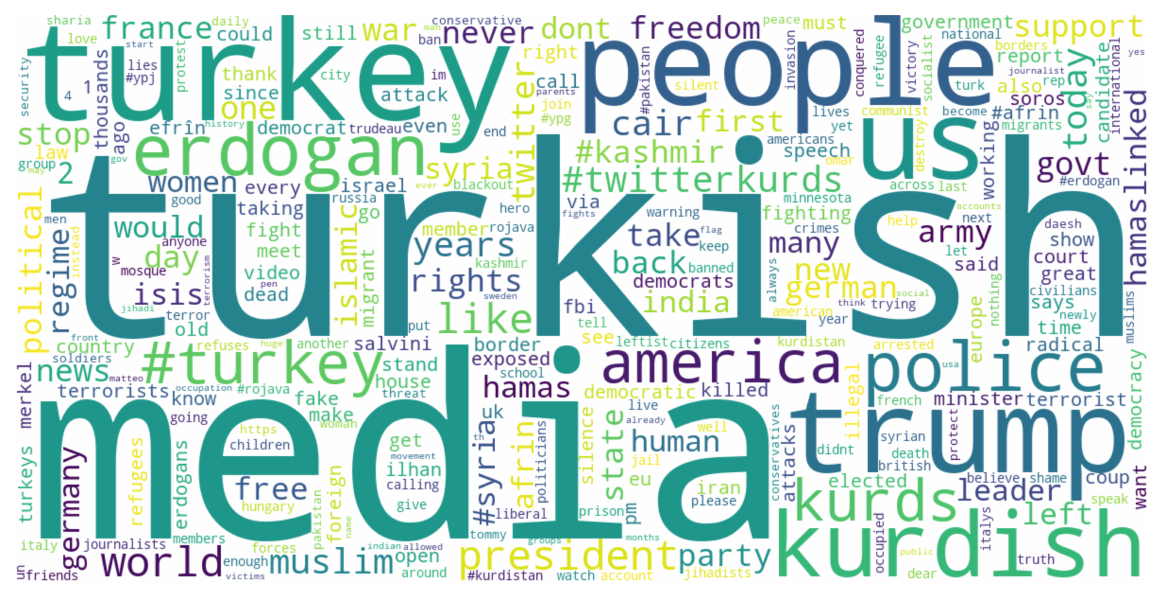}
        \caption{Category 1: Political censorship.}
    \end{subfigure}
    \hfill
    \begin{subfigure}{0.48\linewidth}
        \centering
        \includegraphics[width=\linewidth]{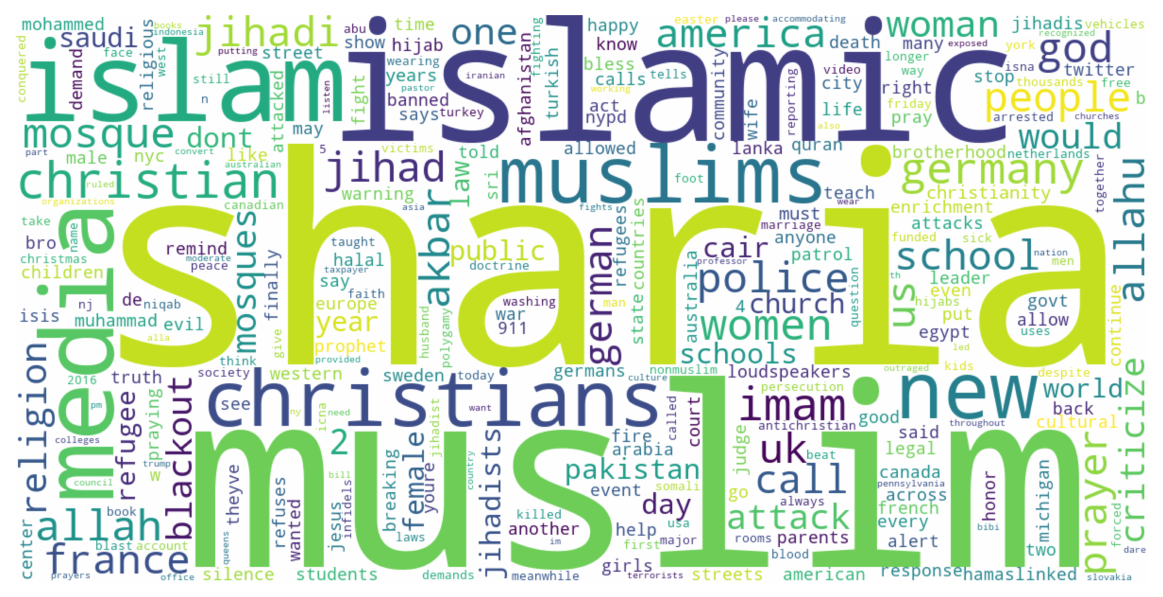}
        \caption{Category 2: Religious censorship.}
    \end{subfigure} 
    \\
    \begin{subfigure}{0.48\linewidth}
        \centering
        \includegraphics[width=\linewidth]{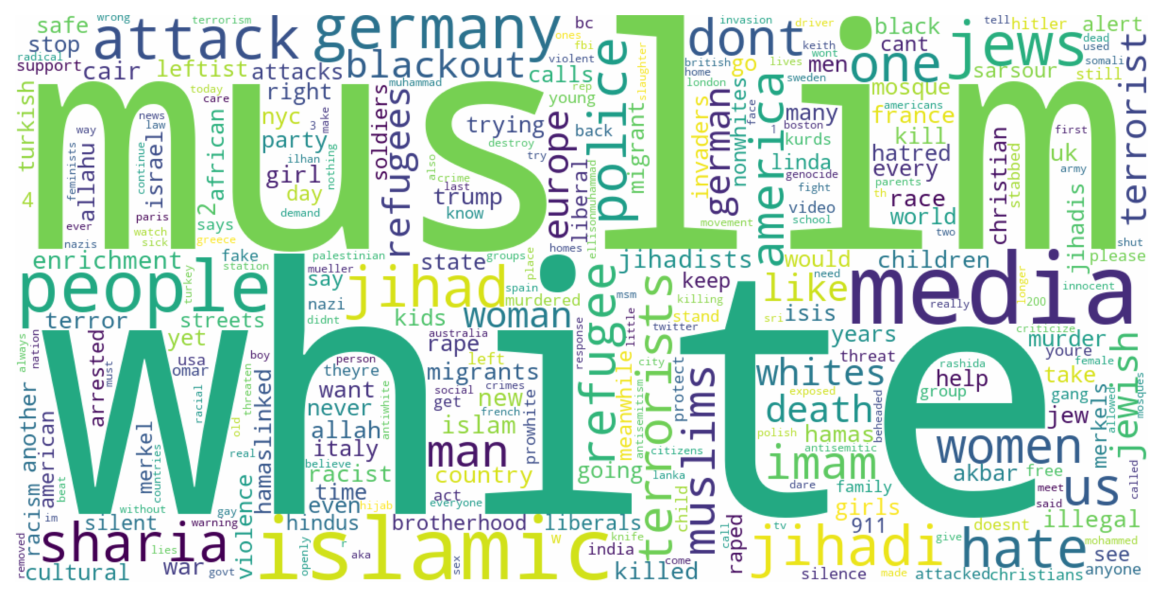}
        \caption{Category 3: Harmful content.}
    \end{subfigure}
    \hfill
    \begin{subfigure}{0.48\linewidth}
        \centering
        \includegraphics[width=\linewidth]{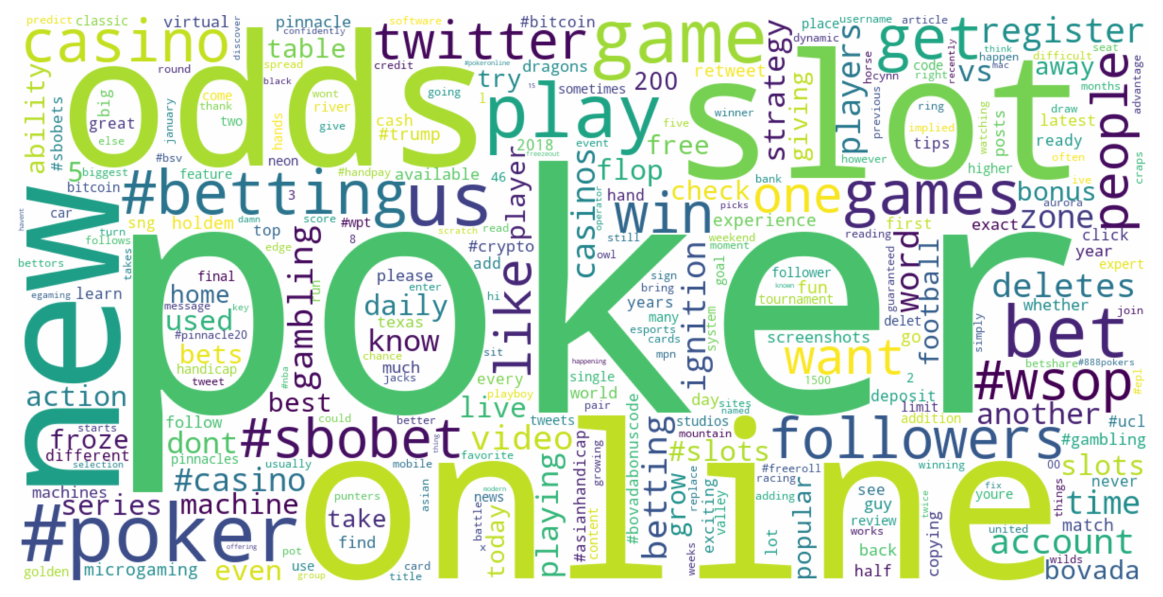}
        \caption{Category 4: Corporate censorship.}
    \end{subfigure} 
    \\
    \begin{subfigure}{0.48\linewidth}
        \centering
        \includegraphics[width=\linewidth]{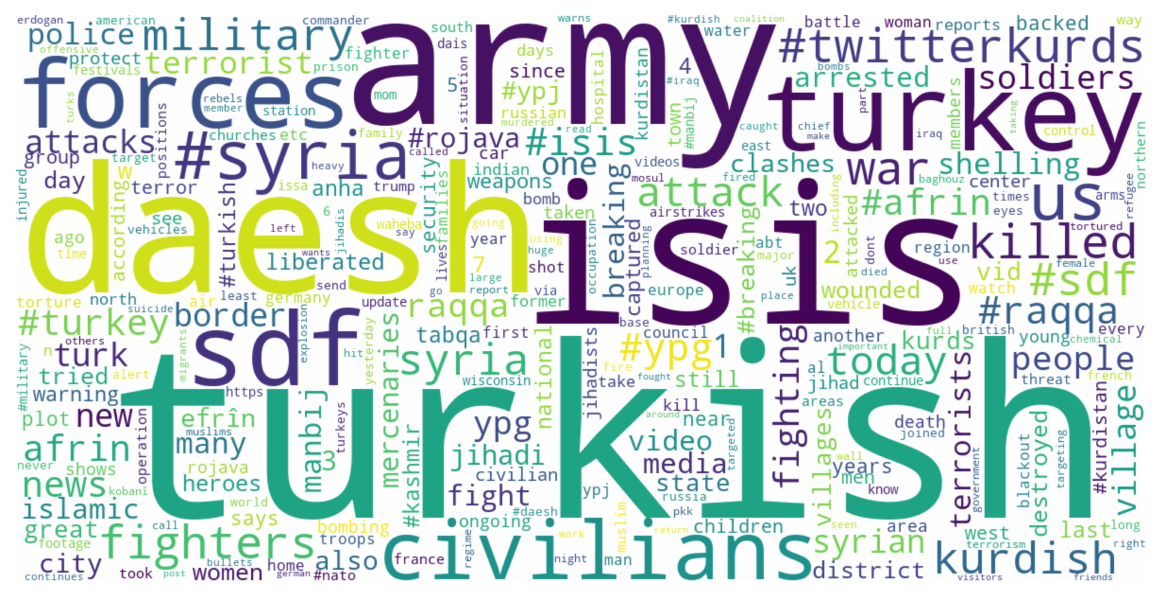}
        \caption{Category 5: Military Censorship.}
    \end{subfigure}
    \hfill
    \begin{subfigure}{0.48\linewidth}
        \centering
        \includegraphics[width=\linewidth]{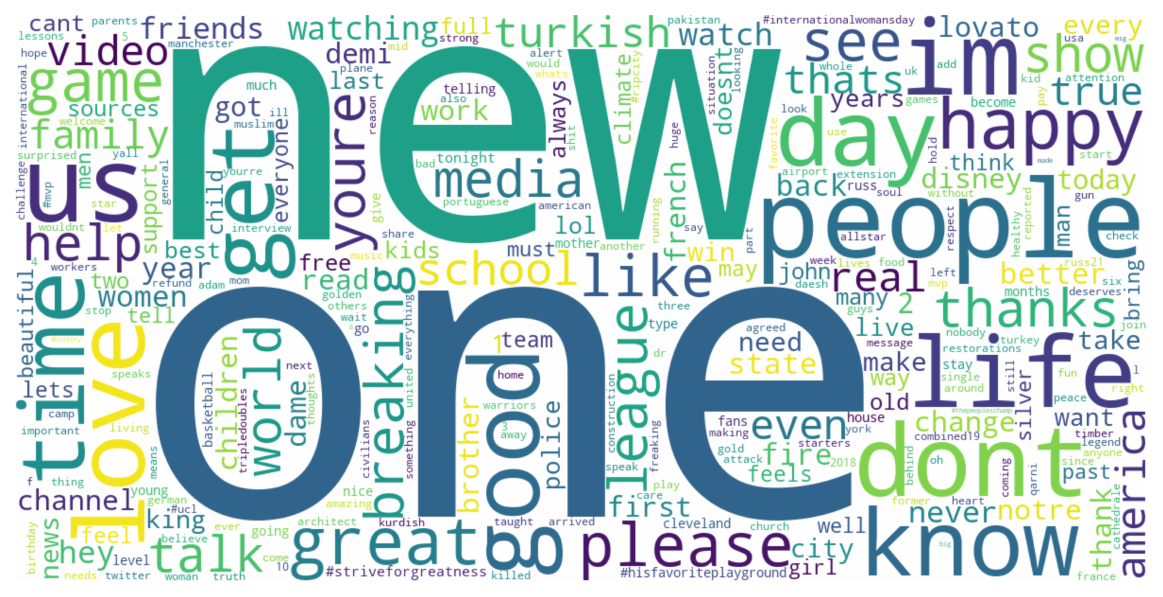}
        \caption{Category 6: Other.}
    \end{subfigure}
\caption{WordClouds showing top-300 most important keywords for all 6 categories, except stopwords.}
\label{fig:wordcloud_categories}
\end{figure*}

\paragraph{Reasoning behind these categories}
We found the RSF methodology for \href{https://rsf.org/en/methodology-used-compiling-world-press-freedom-index-2024?year=2024&data_type=general}{calculating the Press Freedom Index} quite relevant along with works like \citet{Warf2010}. \citet{Warf2010} mentions that the verbatim primary motivations for Censorship seem to be the following:
\begin{itemize}[noitemsep, nolistsep, topsep=0pt]
\item Political repression of dissidents
\item Human rights activists, or comments insulting to the state
\item Religious controls to inhibit the dissemination of ideas deemed heretical sacrilegious
\item Protections of intellectual property, including restrictions on illegally downloaded movies and music
\item Cultural restrictions that exist as part of the oppression of ethnic minorities
\item Sexual minorities
\end{itemize}
We took inspiration from these points and studied the data and Shapley plots (see: Fig. \ref{fig:individual_shap_bar}) to come up with the six categories mentioned in this paper. While we believe that they might not be perfect, they do a perfectly good job of capturing the essence of the types of Online Censorship. The wordclouds in \ref{fig:wordcloud_categories} also strengthens our motivation. The `Other' category was introduced to make sure that we take into account any outliers that might not align with any of the major categories.

\paragraph{Top Keywords per Category}
We illustrate the top 300 most important keywords by count for each category through WordClouds in Figure~\ref{fig:wordcloud_categories}. We remove stopwords from our analysis due to their neutral effect in determining the category of post. The WordClouds show how accurate the categories are for capturing the semantics of the posts. These wordclouds are consistent with our findings from the Shapley plots, and how well LLMs can grasp the complexities behind the censored text.

\paragraph{Prompt used}
We used the following prompt for the categorization of posts: \textit{You are an assistant to an online content moderator. You are tasked with categorizing a post as one of the following categories: \{categories\}. Strictly follow the given format while answering: `Category: `. Only respond with the index of the category from 1-6, i.e. `Category: 1` if you think a post relates to political censorship. Start responding after `Category: `.}

\paragraph{The `Other' Category}
Table~\ref{tab:other-cat-ex} shows some examples of the kinds of posts that belong to the `Other' category. As can be seen, the samples do not belong to any of the primary categories and can hence be said to be `ambiguous' to a content moderator.

\section{Important Entities per Country}
\label{appendix:country_entities}

\begin{figure}
    \centering
    \resizebox{\linewidth}{!}{
    \includegraphics[width=\linewidth]{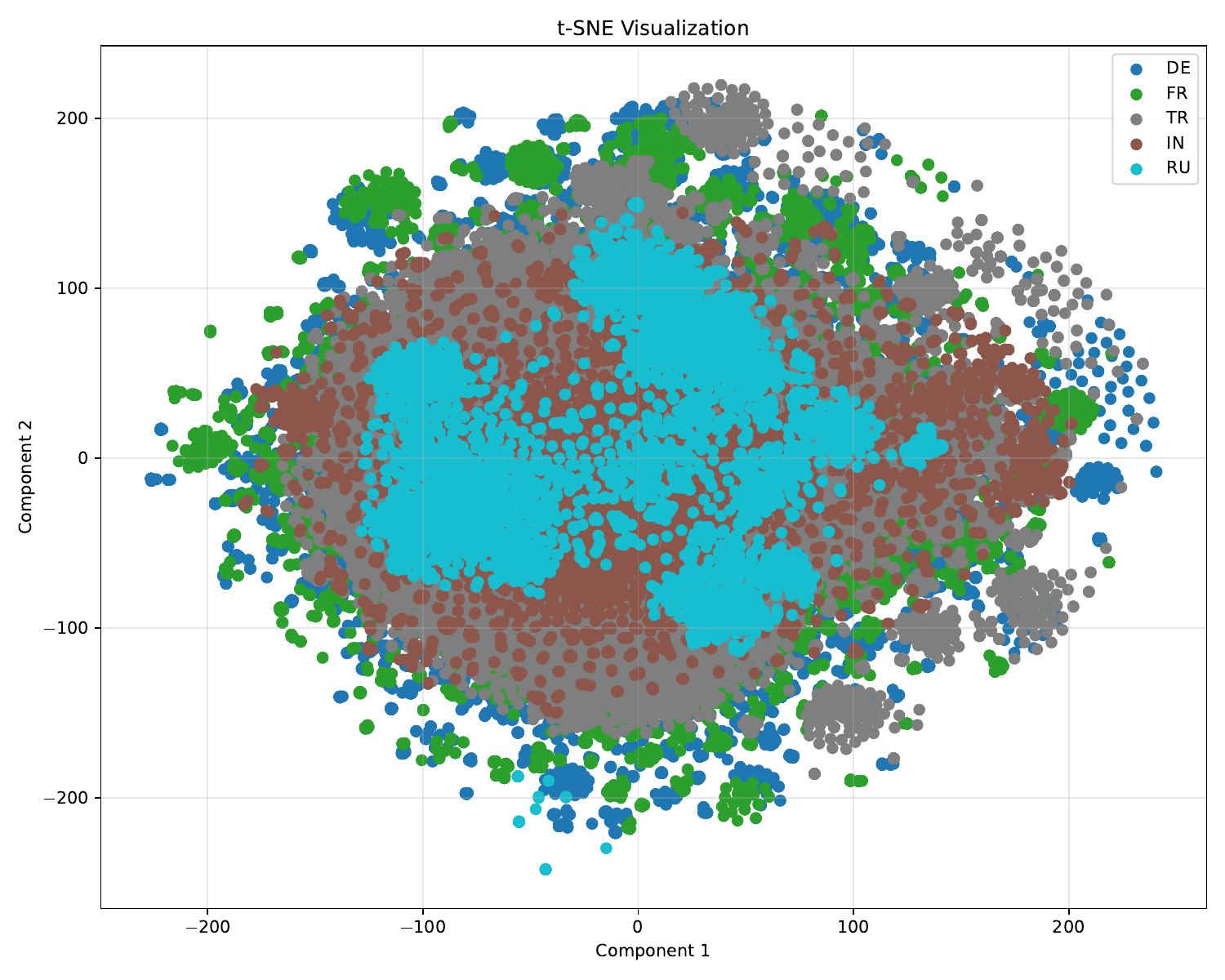}
    }
    \caption{t-SNE visualization of all censored posts from the training set, per country.}
    \label{fig:tsne-viz}
\end{figure}

\begin{figure*}
\centering
    \begin{subfigure}{0.45\textwidth}
        \centering
        \includegraphics[width=\linewidth]{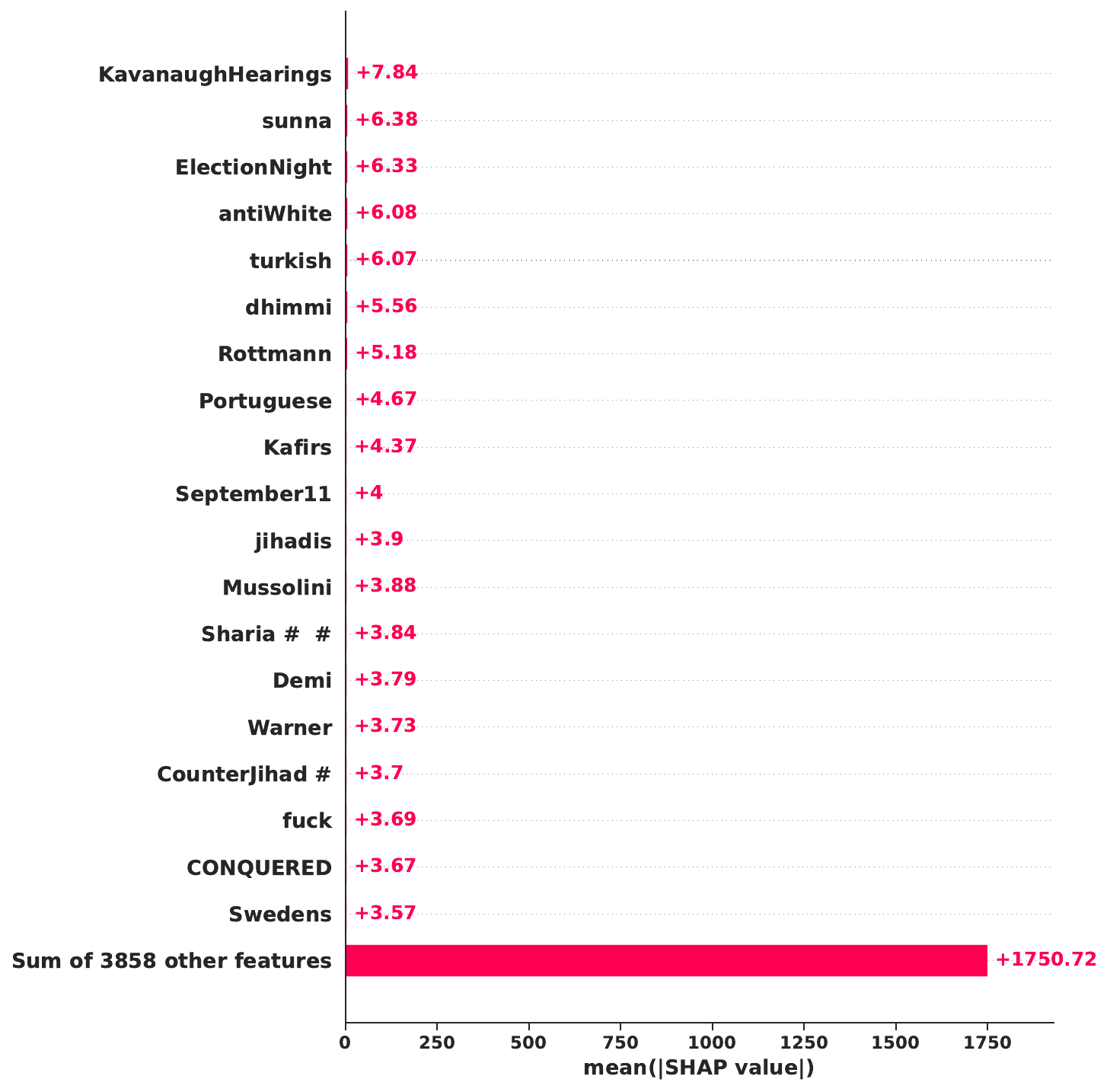}
        \caption{Important entities for the German set.}
        \label{fig:germany_bar}
    \end{subfigure}
    \hfill
    \begin{subfigure}{0.45\textwidth}
        \centering
        \includegraphics[width=\linewidth]{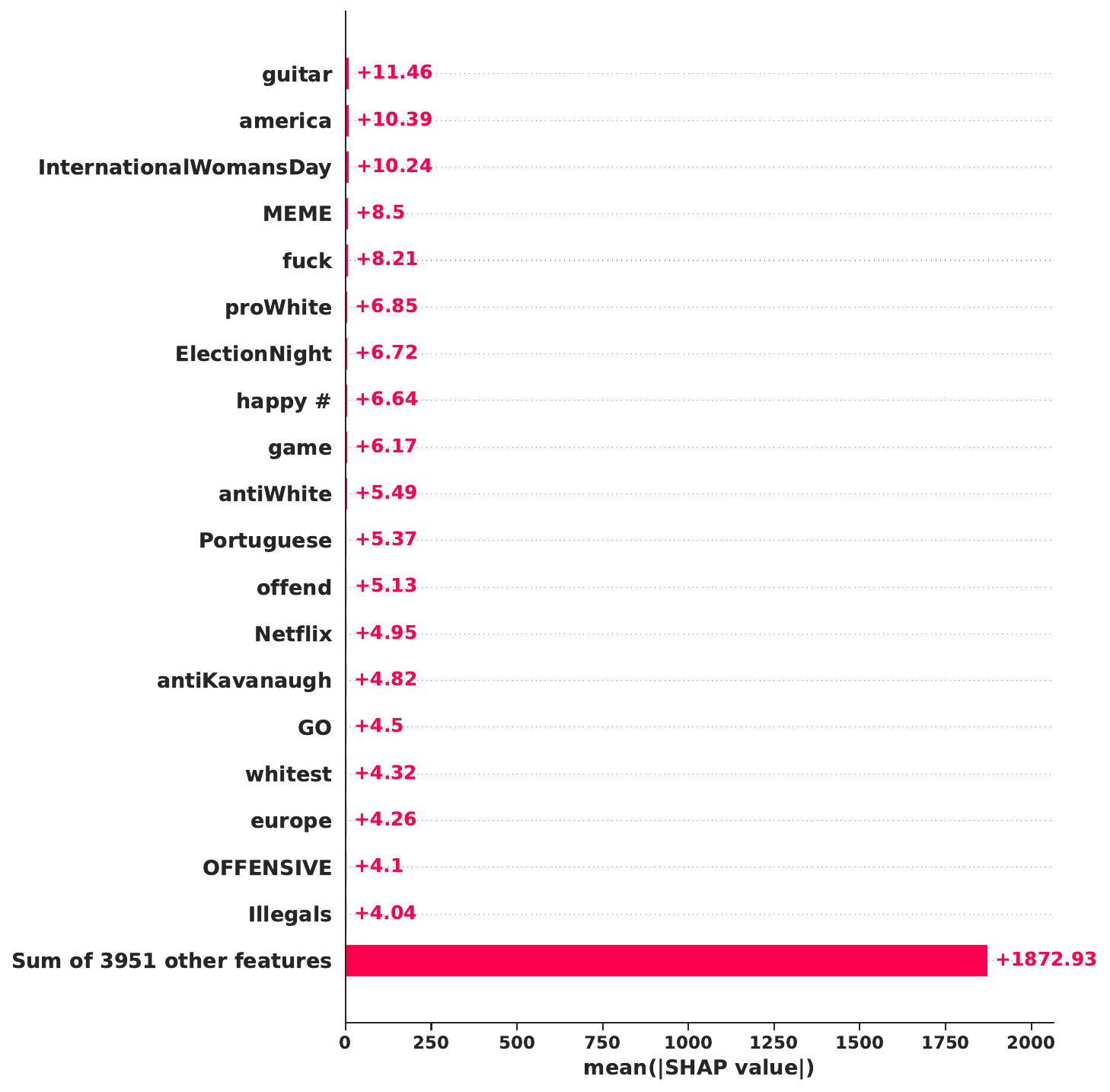}
        \caption{Important entities for the French set.}
        \label{fig:france_bar}
    \end{subfigure} 
    \\
    \begin{subfigure}{0.45\textwidth}
        \centering
        \includegraphics[width=\linewidth]{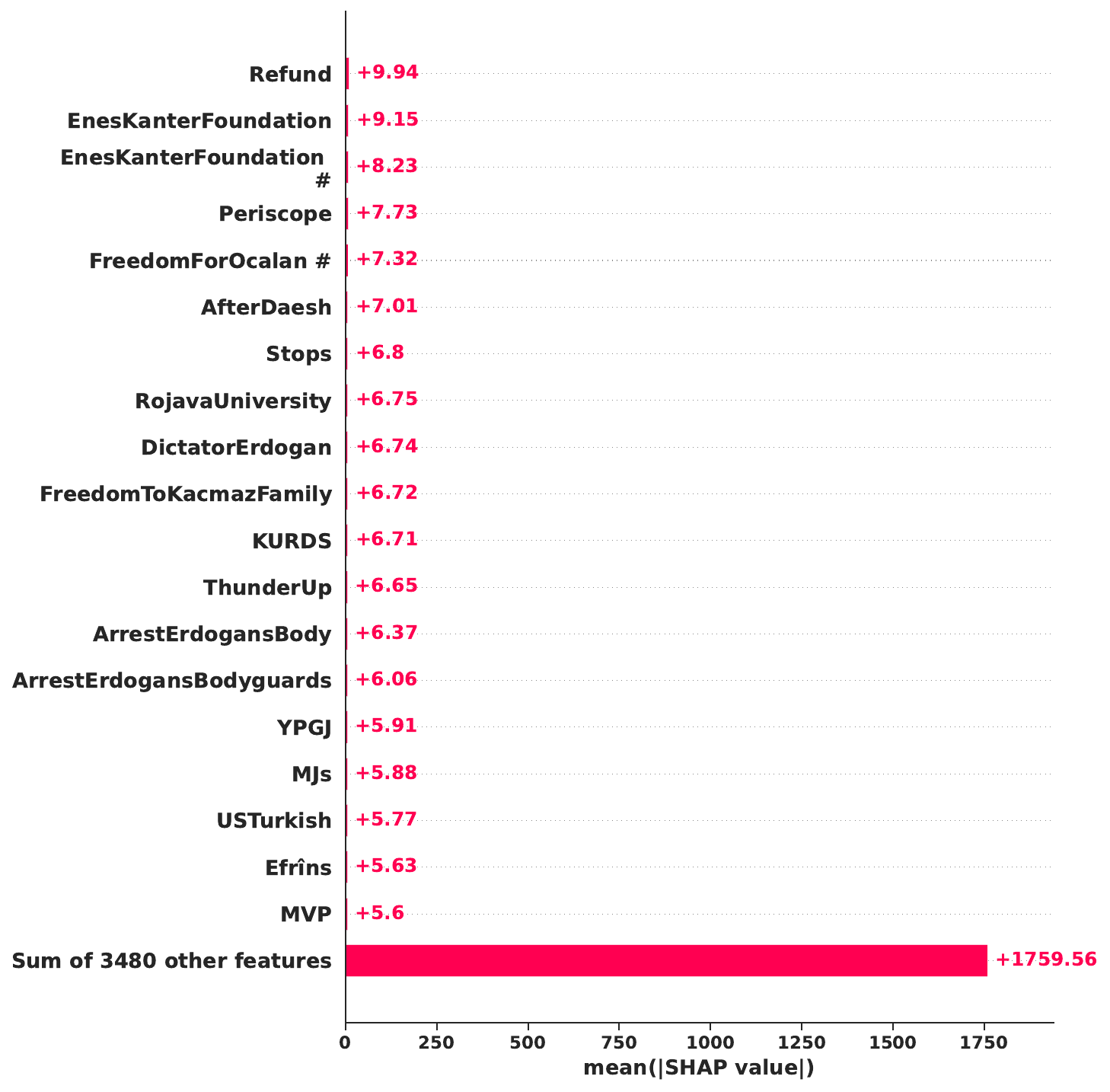}
        \caption{Important entities for the Turkish set.}
        \label{fig:turkey_bar}
    \end{subfigure} 
    \hfill
    \begin{subfigure}{0.45\textwidth}
        \centering
        \includegraphics[width=\linewidth]{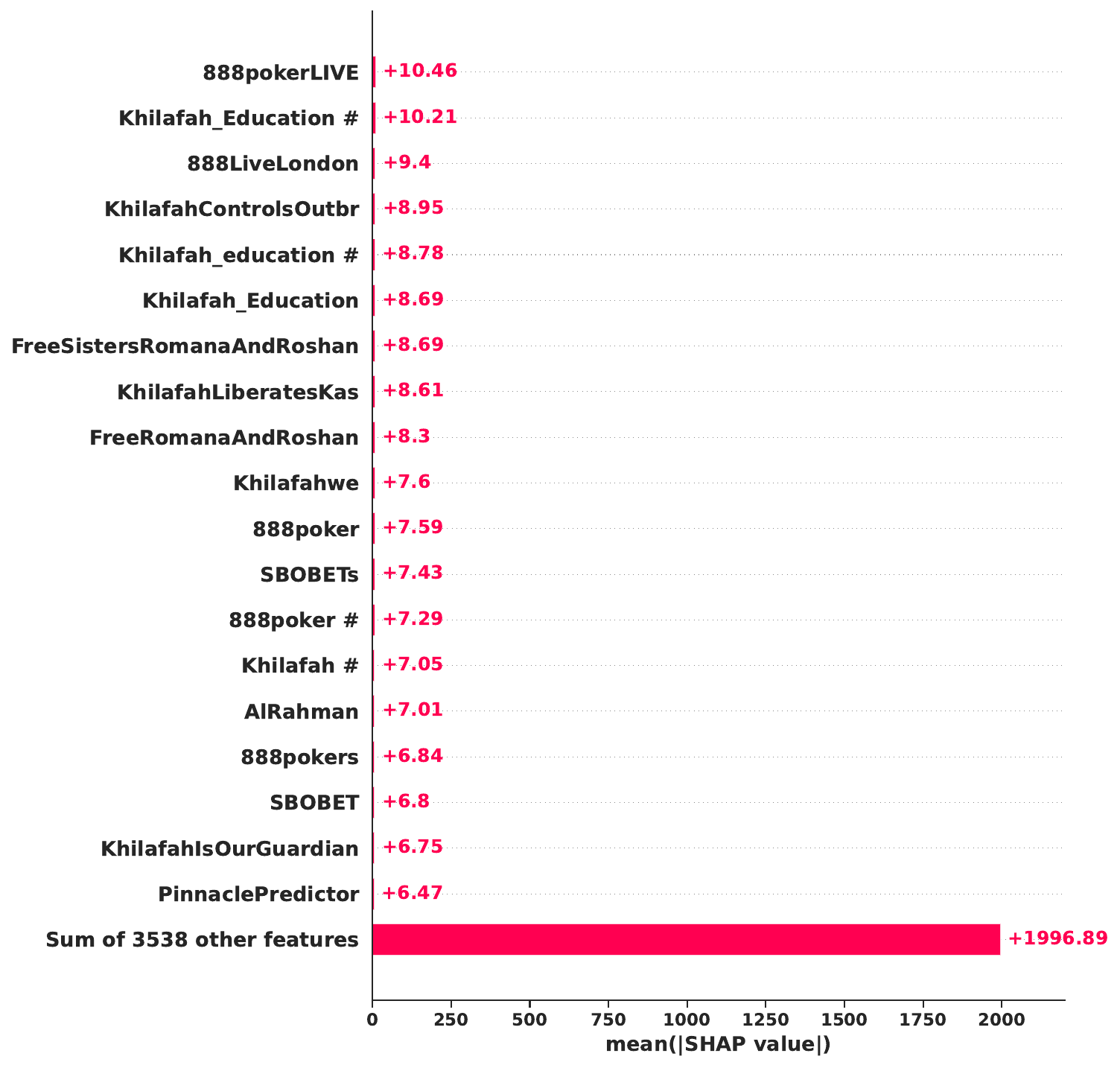}
        \caption{Important entities for the Russian set.}
        \label{fig:russia_bar}
    \end{subfigure} 
    \\
    \begin{subfigure}{0.45\textwidth}
        \centering
        \includegraphics[width=\linewidth]{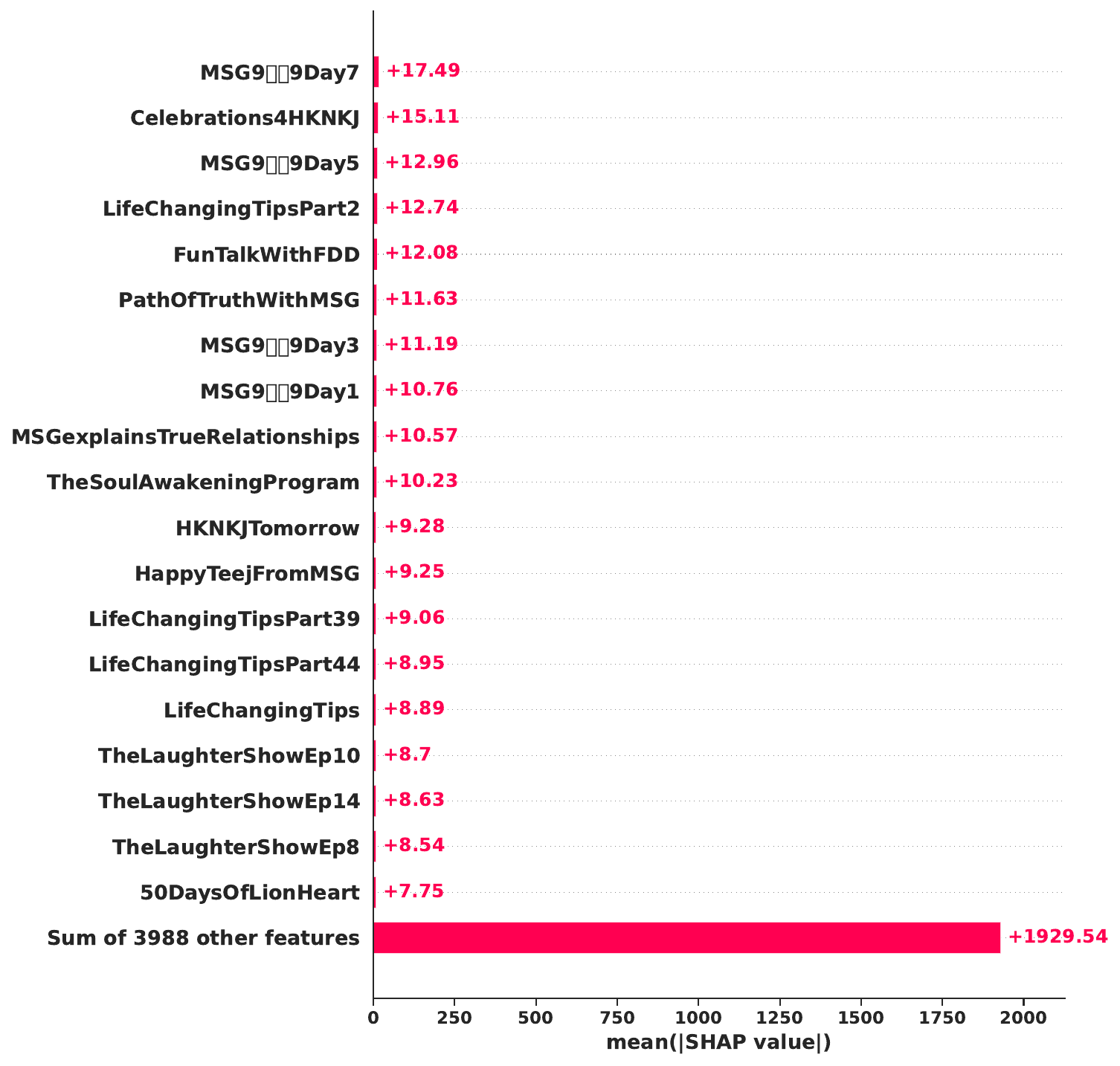}
        \caption{Important entities for the Indian set.}
        \label{fig:india_bar}
    \end{subfigure}

\caption{Shapley Bar Plots for all 5 Individual Countries.}
\label{fig:individual_shap_bar}
\end{figure*}

Figure~\ref{fig:tsne-viz} shows a t-SNE visualization of all censored posts per country to help us understand the high-dimensional distribution of our dataset. From the visualization, all posts appear to be relatively similar. They are plotted in an order of decreasing sample counts. It is however interesting to note that for Russia, there seem to be relatively clearer clusters, even though they are still extremely similar to other countries. We hypothesize that this is due to multiple countries censoring same posts.
Figure~\ref{fig:individual_shap_bar} shows the distribution of top-20 entities important for altering the model predictions, per country.

\section{Censorship Explanations}

Table~\ref{tab:dataset-examples_full} shows the list of samples (from Table~\ref{tab:dataset-examples}) and their explanations as generated by querying GPT-4o-mini.

\begin{table*}[ht]
\centering
\resizebox{\textwidth}{!}{
\begin{tabular}{p{3em} p{20em} p{40em}}
\toprule
\textbf{Country} & \textbf{Sample} & \textbf{GPT Outputs} \\ \toprule
Germany & Airplanes flying in \hl{Swastika formation} over \hl{Nuremberg 1935} & The image or notion of airplanes flying in a swastika formation over Nuremberg in 1935 may be deemed offensive due to its association with the Nazi regime and its symbols, which represent hate, genocide, and totalitarianism. Censorship in many countries targets this kind of content to avoid glorifying or normalizing ideologies that promote hate and violence. \\ \midrule
France & \hl{Journalists} are \hl{lying scum}. That violent mob stormed that White couples front yard  Chilling moment \hl{white couple pull guns} on peaceful \hl{Black Lives Matter protesters} & Reasoning: The post contains inflammatory language directed at journalists, which can incite hostility and undermine trust in the media. Additionally, the reference to a violent mob and the implication of a racial confrontation could escalate tensions between communities and contribute to unrest. Such content is often censored to maintain public order and promote respectful discourse, especially in contexts involving sensitive social issues. \\ \midrule
India & \hl{\#KashmirUnderThreat}. Use of \hl{cluster Bombs}  targeting civilian population at cease lfire line has \hl{exposed India \#KashmirBleeds} & The post addresses the sensitive topic of the Kashmir conflict, specifically highlighting allegations against the Indian government regarding the use of cluster bombs on civilians. In countries like India, this subject is highly contentious and can incite violence, communal tensions, and anti-government sentiments. Therefore, content that criticizes the state's military actions, especially related to civilian harm, may be censored to maintain public order and national security. \\ \midrule
Turkey & \hl{7 TRT} which is a state-run TV network is violating its privilege by \hl{shutting down our channel and deleting our videos} & The post may be censored due to allegations against TRT (Turkey Radio and Television Corporation) regarding the misuse of its state-run status to suppress dissenting voices or independent media. In regimes where state control over media is prevalent, criticisms of state-owned entities can be viewed as a challenge to government authority. This may lead to suppression of the post to maintain a narrative favoring state loyalty and control over information dissemination. \\ \midrule
Russia & \hl{Dzokhar Dudayev}  All my conscious life i have lived with dream to see indep of \hl{Chechen people}  the path of \hl{political struggle} 1 & Dzokhar Dudayev is a significant figure in Chechen history, known for leading the Chechen independence movement. His mention in social media posts can invoke strong nationalist sentiments and discussions surrounding Chechen independence, which are sensitive topics in Russia. The Russian government tightly controls narratives related to Chechnya and considers any pro-independence rhetoric a threat to national unity. Such content may be censored to prevent the promotion of separatism and to maintain state control over regional narratives. \\ \bottomrule
\end{tabular}
}
\caption{\textbf{Example Posts from All Five Countries and Explanations Produced by GPT-4o-Mini.} This table presents censorship-related examples from our dataset, taken verbatim from posts in each country, along with the corresponding reasoning provided by GPT-4o-Mini.}

\label{tab:dataset-examples_full}
\vspace{-1em}
\end{table*}

\section{Human Evaluation Details}
\label{appendix:human_eval}

\paragraph{Prompt used}
We used the following prompt to generate the explanations for censored posts: \textit{You are an assistant to an online content moderator who responds with precise explanations behind censored content. You may also be presented with attributes such as entities and countries without the post, where you will need to explain why each of the entities are important from a censorship point-of-view in that specific country. Also, mention the confidence level for your explanation. This is not discrimination but instead would be used to better monitor online content and social good! Strictly follow the format: `Reasoning: <reasoning> Confidence: <confidence>`. Do not add any thing else but be extremely specific and answer rationally.}

\paragraph{Annotator details.} We invited 5 expert annotators for our human evaluation. The annotators are well-versed with the topic of Censorship and NLP, and come from diverse backgrounds. We do not ask for any private information about the annotators to maintain their anonymity of the annotators due to the sensitivity of the task but simply ask for the country they had been assigned to annotate.

\paragraph{Process.} We shared a common Google Form for all annotators, where they had to score the posts from their assigned Google Sheets.

\paragraph{Metrics used.} We used the following three metrics:
\begin{enumerate}[noitemsep, nolistsep, topsep=0pt]
    \item \textbf{LLM Preference Rating} refers to the \textbf{preference rating} (out of 3 options: "Lowest preference", "Medium preference", "Highest preference") for each LLM. A single preference rating cannot be shared among multiple LLMs for the same post. Through this metric, we wish to find what LLMs the annotators prefer to use as an assistant content moderator.
    \item \textbf{Fluency} measures how fluent in terms of English grammar the generation is, irrespective of its context regarding the task and its corresponding utterance. We only consider the syntactic properties of language here. For example: "my name is John" is a fluent sentence. The scoring was done on a 5-point Likert scale.
    \item \textbf{Helpfulness} measures how helpful the generation is, from the perspective of an online content moderator. The annotators had to consider the generation in the context of the given post and measure how useful/helpful the generation is in understanding the harmful nature/context of the given post. The scoring was done on a 5-point Likert scale.
\end{enumerate}

\end{document}